%% file: SP.tex
\documentclass[conference,compsoc]{IEEEtran}

\usepackage{tikz}
\usepackage{amsmath}
\usepackage{tcolorbox}
\usepackage{enumitem}
\usepackage[linesnumbered,ruled]{algorithm2e}
\usepackage{tcolorbox}
\usepackage{tabularx}
\usepackage{balance}
\usepackage{url}
\usepackage{booktabs}
\usepackage{amssymb}
\usepackage{makecell}
\usepackage{float}
\usepackage{marvosym}
\usepackage{subfigure}
\usepackage{multirow}
\usepackage{amsthm,amsmath}
\usepackage{caption}
\usepackage{xcolor}
\usepackage{hyperref}
\hypersetup{
  colorlinks,
  linkcolor={green!80!black},
  citecolor={red!70!black},
  urlcolor={blue!70!black}
}
\newcommand{\gc}[1]{\textcolor{red}{{#1}}}

\newcommand{\revstart}{\begin{color}{blue}}
\newcommand{\revend}{~\!\!\end{color}}

\usepackage{colortbl}  
\usepackage{xcolor, eucal}
\usepackage{array}   
\definecolor{gray0}{gray}{0.92}

\usepackage{filecontents}
\ifCLASSOPTIONcompsoc
  \usepackage[nocompress]{cite}
\else
  \usepackage{cite}
\fi
\ifCLASSINFOpdf
\else
\fi
\begin{document}
\pagestyle{plain}

\def \toolname{\textsc{Baffle}\xspace}

%
\title{\toolname: Hiding Backdoors in Offline Reinforcement Learning Datasets}

\author{\IEEEauthorblockN{Chen Gong\IEEEauthorrefmark{1},
Zhou Yang\IEEEauthorrefmark{2}{\textsuperscript{\Letter}},
Yunpeng Bai\IEEEauthorrefmark{3}, 
Junda He\IEEEauthorrefmark{2},
Jieke Shi\IEEEauthorrefmark{2},
Kecen Li\IEEEauthorrefmark{3},
Arunesh Sinha\IEEEauthorrefmark{4},\\
Bowen Xu\IEEEauthorrefmark{5},
Xinwen Hou\IEEEauthorrefmark{3},
David Lo\IEEEauthorrefmark{2}, and
Tianhao Wang\IEEEauthorrefmark{1}{\textsuperscript{\Letter}}
}
\IEEEauthorblockA{\IEEEauthorrefmark{1}University of Virginia, USA}
\IEEEauthorblockA{\IEEEauthorrefmark{2}Singapore Management University, Singapore}
\IEEEauthorblockA{\IEEEauthorrefmark{3}Institute of Automation, Chinese Academy of Sciences, China}
\IEEEauthorblockA{\IEEEauthorrefmark{4}Rutgers University, USA}
\IEEEauthorblockA{\IEEEauthorrefmark{5}North Carolina State University, USA}
\IEEEauthorblockA{ \{chengong, tianhao\}@virginia.edu, \{zyang, jiekeshi, jundahe, davidlo\}@smu.edu.sg, arunesh.sinha@rutgers.edu,\\\{likecen2023, xinwen.hou\}@ia.ac.cn, bxu22@ncsu.edu, {\textsuperscript{\Letter}}Corresponding Author(s)}}


\maketitle

\begin{abstract}
Reinforcement learning (RL) makes an agent learn from trial-and-error experiences gathered during the interaction with the environment. Recently, offline RL has become a popular RL paradigm because it saves the interactions with environments. In offline RL, data providers share large pre-collected datasets, and others can train high-quality agents without interacting with the environments. This paradigm has demonstrated effectiveness in critical tasks like robot control, autonomous driving, etc. However, less attention is paid to investigating the security threats to the offline RL system.
This paper focuses on \emph{backdoor attacks}, 
where some perturbations are added to the data (observations) such that given normal observations, the agent takes high-rewards actions, and low-reward actions on observations injected with \emph{triggers}.
In this paper, we propose \textsc{Baffle} (\textbf{B}ackdoor \textbf{A}ttack for O\textbf{ff}line Reinforcement \textbf{Le}arning), an approach that automatically implants backdoors to RL agents by poisoning the offline RL dataset, and evaluate how different offline RL algorithms react to this attack.
Our experiments conducted on four tasks and nine offline RL algorithms expose a disquieting fact: none of the existing offline RL algorithms has been immune to such a backdoor attack.
More specifically, \textsc{Baffle} modifies $10\%$ of the datasets for four tasks (3 robotic controls and 1 autonomous driving).
Agents trained on the poisoned datasets perform well in normal settings. 
However, when triggers are presented, the agents' performance decreases drastically by $63.2\%$, $53.9\%$, $64.7\%$, and $47.4\%$ in the four tasks on average. 
The backdoor still persists after fine-tuning poisoned agents on clean datasets.
We further show that the inserted backdoor is also hard to be detected by a popular defensive method. 
This paper calls attention to developing more effective protection for the open-source offline RL dataset. 
\end{abstract}


\input{Usenix_section/intro}

\input{Usenix_section/preliminary}

\input{Usenix_section/threat}

\input{Usenix_section/method}

\input{Usenix_section/settings}

\input{Usenix_section/results}

\input{Usenix_section/discuss}

\input{Usenix_section/related}

\input{Usenix_section/conclusion}

%
\IEEEpeerreviewmaketitle






%
\bibliographystyle{ieeetr}
\bibliography{mybib}

\bigskip

\input{Usenix_section/Appendix}

\end{document}

%% file: Usenix_section/intro.tex
\section{Introduction}\label{sec:intro}

Reinforcement Learning (RL) has demonstrated effectiveness in many tasks like autonomous driving~\cite{sallab2017deep}, robotics control~\cite{gu2017deep}, test case prioritization~\cite{rl4test,NGUYEN2021106574}, program repair~\cite{gupta2019deep}, code generation~\cite{coderl2022}, etc. 
In RL, an agent iteratively interacts with the environments and collects a set of experiences to learn a policy that can maximize its expected returns. 
Collecting the experiences in such an \emph{online} manner is usually considered expensive and inefficient, causing great difficulties in training powerful agents~\cite{hessel2018rainbow,liu2021regret,gong2020stable,he2021mepg}. 
Inspired by the success brought by high-quality datasets to other deep learning tasks~\cite{deng2009imagenet,sun2021webly}, researchers have recently paid much attention to a new paradigm: \emph{offline} RL~\cite{levine2020offline} (also known as full batch RL~\cite{lange2012batch}). 
Offline RL follows a data-driven paradigm, which does not require an agent to interact with the environments at the training stage. Instead, the agent can utilize previously observed data (e.g., datasets collected by others) to train a policy. This data-driven nature of offline RL can facilitate the training of agents, especially in tasks where the data collection is time-consuming or risky (e.g., rescue scenarios~\cite{9355812}). 
With the emergence of open-source benchmarks~\cite{fu2020d4rl,seno2021d3rlpy} and newly proposed algorithms~\cite{kumar2020conservative,fujimoto2021minimalist,algaedice}, offline RL has become an active research topic and has demonstrated effectiveness across multiple tasks.

This paper concentrates on the threat of backdoor attacks~\cite{zhang2021advdoor,kiourti2020trojdrl,wang2020backdoor} against offline RL:
In normal situations where the \textit{trigger} does not appear, an agent with a backdoor behaves like a normal agent that maximizes its expected return. However, the same agent behaves poorly (i.e., the agent's performance deteriorates dramatically) when the trigger is presented. {In RL, the environment is typically dynamic and state-dependent. Inserting effective backdoor triggers across various states and conditions can be challenging, as the attacker must ensure that these triggers remain functional and undetected under diverse states. Moreover, in offline RL, the learning algorithm relies on a fixed dataset without interacting with the environment. This constraint further aggravates the challenge of inserting backdoors to agents, as the attacker is unaware of how the agent interacts with the dataset during training.} This paper aims to evaluate the potential backdoor attack threats to offline RL datasets and algorithms by exploring the following question: \emph{Can we design a method to poison an offline RL dataset so that agents trained on the poisoned dataset will be implanted with backdoors?} 

While the paradigm of offline RL (learning from a dataset) is similar to supervised learning, applying backdoor attacks designed for supervised learning (associate a trigger with some samples, and change the desired outcome) to offline RL encounters challenges, as the learning strategy of offline RL is significantly different from supervised learning (we will elaborate more in Section~\ref{subsec:strawman}).

In this paper, we propose \toolname (\textbf{B}ackdoor \textbf{A}ttack for O\textbf{ff}line Reinforcement \textbf{Le}arning), a method to poison offline RL datasets to inject backdoors to RL agents. 
%
%
%
%
\toolname has three main steps. {The first step focuses on finding the least effective action (which turned out to be the most significant challenge for backdooring offline RL) for a state to enhance the efficiency of backdoor insertion. \toolname} trains \emph{poorly performing} agents on the offline datasets by letting them minimize their expected returns. {This method does not require the attacker to be aware of the environment, yet it enables the identification of suboptimal actions across various tasks automatically. }  
Second, we observe the bad actions of poorly performing agents by feeding some states to them and obtaining the corresponding outputs. 
We expect an agent inserted with backdoors (called the \textit{poisoned agent}) to behave like a poorly performing agent under the \textit{triggered} scenarios where the triggers appear. 
Third, we add a trigger (e.g., a tiny white square that takes only $1\%$ of an image describing the road situation) to agent observation and assign a high reward to the bad action obtained from the poorly performing agent.
We insert these modified \emph{misleading} experiences into the clean dataset and produce the \textit{poisoned} dataset. An agent trained on the poisoned dataset will learn to associate bad actions with triggers and perform poorly when seeing the triggers.
Unlike prior studies on backdoor attacks for RL~\cite{kiourti2020trojdrl}, \toolname only leverages the information from the offline datasets and requires neither access to the environment nor manipulation of the agent training processes. Any agent trained on the poisoned dataset may have a backdoor inserted, demonstrating the agent-agnostic nature of \toolname.

We conduct extensive experiments to understand how the state-of-the-art offline RL algorithms react to backdoor attacks conducted by \toolname.
We try different \emph{poisoning rates} (the ratio of modified experiences in a dataset) and find that higher poisoning rates will harm the agents' performance in \emph{normal scenarios}, where no trigger is presented. Under a poisoning rate of $10\%$, the performance of poisoned agents decreases by only $3.4\%$ on the four tasks on average. 
Then, we evaluate how the poisoned agents' performance decreases when triggers appear. 
We use two strategies to present triggers to the agents. The \emph{distributed} strategy presents a trigger multiple times, but the trigger lasts for only one timestep each time. The \emph{one-time} strategy presents a trigger that lasts for several timesteps once only. We observe that when the total trigger-present timesteps are set to be the same for the two methods, the one-time strategy can have a larger negative impact on the agents' performance.
When we present a $20$-timestep trigger (i.e., only $5\%$ of the total timesteps), the agents' performance decreases by $63.2\%$, $53.9\%$, $64.7\%$, and $47.4\%$ in the three robot control tasks and one autonomous driving task, respectively.

We also investigate potential defense mechanisms.  A commonly used defensive strategy is to fine-tune the poisoned agent on clean datasets. Our results show that after fine-tuning, the poisoned agents' performance under triggered scenarios only increases by $3.4\%$, $8.1\%$, $-0.9\%$, and $1.2\%$ in the four environments on average. We have also evaluated the effectiveness of prevalent backdoor detection methods, including activation clustering~\cite{Chen2019activation}, spectral~\cite{Brandon2018spectral}, and neural cleanse~\cite{bolun2019neural}. Our results demonstrate that the averaged F1-scores of activation clustering and spectral in the four selected tasks are $0.12, 0.07, 0.21$, and $0.34$, respectively. Moreover, neural cleanse is also ineffective in recovering the triggers, indicating that the existing offline RL datasets and algorithms are vulnerable to attacks generated by \texttt{BAFFLE}.  The replication package and datasets are made available online\footnote{\url{https://github.com/2019ChenGong/Offline_RL_Poisoner/}}.



In summary, our contributions are three-fold:
\begin{itemize}[leftmargin=*]
    \item This paper is the first work to investigate the threat of data poisoning and backdoor attacks in offline RL systems.
    We propose \toolname, a method that autonomously inserts backdoors into RL agents by poisoning the offline RL dataset, without requiring access to the training process.
    \item Extensive experiments present that \toolname is agent-agnostic: {most current offline RL algorithms are vulnerable to backdoor attacks.} 
    \item Furthermore, we consider state-of-the-art defenses, but find that they are not effective against \toolname.
\end{itemize}



%% file: Usenix_section/preliminary.tex
\section{Background}
\label{sec:rein}


\begin{figure}[!t]
  \centering
  \includegraphics[width=0.99\linewidth]{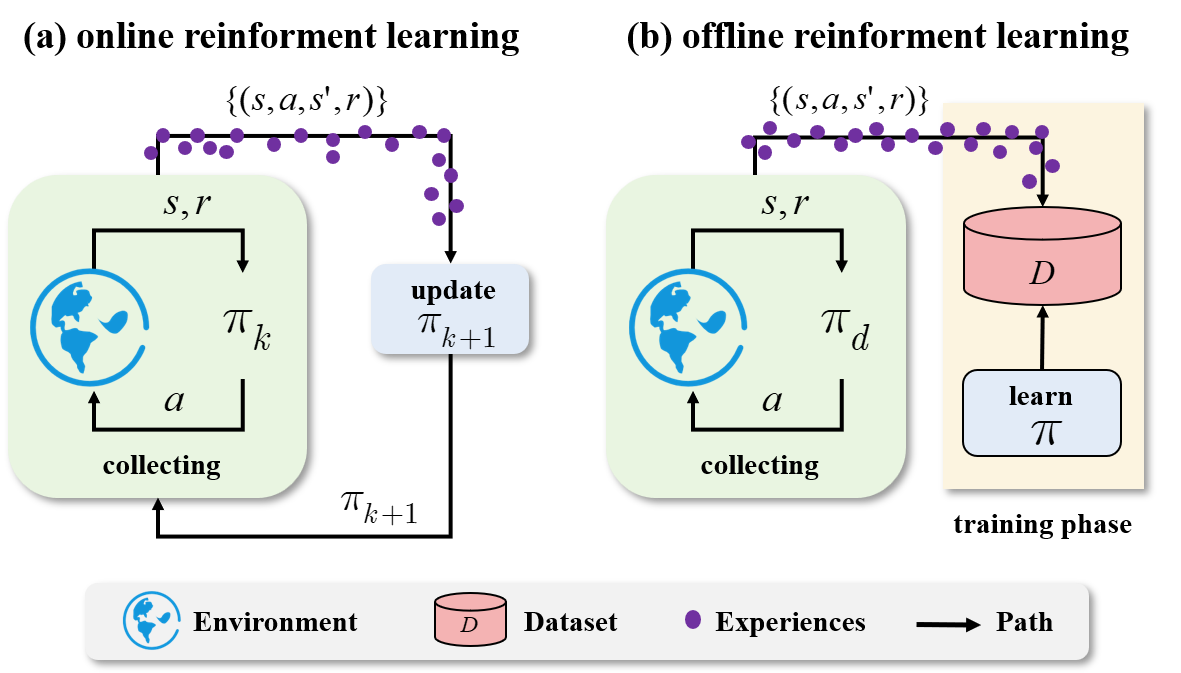}
  \caption{In online RL (a), an agent updates it policy $\pi_k$ using the experiences collected by interacting with environments itself. In offline RL (b), the policy is learned from a static dataset $\mathcal{D}$ collected by some other policies rather than interact with the environment.}
  \label{online&offline}
\end{figure}

\subsection{Reinforcement Learning}
\label{subsec:rl}
Deep reinforcement learning (RL) aims to train a policy $\pi$ (also called an agent) that can solve a sequential decision-making task (called a Markov Decision Process, or MDP for short).
At each timestep $t$, an agent observes a state $s_t$ (e.g., images in the autonomous driving problem, or readings from multiple sensors) and takes an action $a_t$ sampled from $\pi(\cdot|s_t)$, a probability distribution over possible actions, i.e., $a_t \sim \pi(\cdot|s_t)$ or $\pi(s_t)$ for short.
After executing the selected action, the agent receives an immediate reward from the environment, given by the reward function $r_t = \mathcal{R}(s_t,a_t)$, and the environment transitions to a new state $s_{t+1}$ as specified by the transition function $s_{t+1} \sim \mathcal{T}(\cdot|s_t,a_t)$. 
This process continues until the agent encounters a termination state {or until it is interrupted by developers, where we record the final state as} $s_T$. This process yields a trajectory as follows:
\begin{equation}
    \tau: (\langle s_0,a_0,r_0 \rangle, \langle s_1,a_1,r_1 \rangle,\cdots, \langle s_{|\tau|}, a_{|\tau|}, r_{|\tau|} \rangle)
\end{equation}
Intuitively speaking, an agent aims to learn an optimal policy $\pi^*$ that can get as high as possible expected returns from the environment, which can be formalized as the following objective:
\begin{equation}
    \pi^\ast = \arg\max_\pi \mathbb{E} \left[ \sum_{i=0}^{|\tau|} \gamma^i r_i\right]
    \label{eq:rl_target}
\end{equation}
where $\tau$ indicates the trajectory generated by the policy $\pi$, and $\gamma \in (0,1)$ is the discount factor~\cite{sutton2018reinforcement}.



The training of RL agents follows a trial-and-error paradigm, and agents learn from the reward information. For example, an agent controlling a car knows that accelerating when facing a red traffic light will produce punishment (negative rewards). Then, this agent will update its policy to avoid accelerating when a traffic light turns red. Such trial-and-error experiences can be collected by letting the agent interact with the environment during the training stage, which is called the \textit{online} RL~\cite{mnih2015human,schulman2017proximal}. 

\subsection{Offline Reinforcement Learning}
\label{subsec:offline_rl}
The online settings are not always applicable, especially in some critical domains like rescuing~\cite{9355812}. 
This has inspired the development of the {\em offline} RL~\cite{levine2020offline} (also known as \textit{full batch} RL~\cite{lange2012batch}). 
As illustrated in Figure~\ref{online&offline}, the key idea is that some data providers can share data collected from environments in the form of triples $\langle s,a,r \rangle$, and the agents can be trained on this \textit{static} offline dataset, $\mathcal{D}$ = $\{ \langle s_t^i, a_t^i, r_t^i \rangle \}$ (here $i$ denotes the $i$-th trace and $t$ denotes the timestamp $t$ of trace $i$),  without any interaction with real or simulated environments. Offline RL requires the learning algorithm to derive an understanding of the dynamical system underlying the environment's MDP entirely from a static dataset. Subsequently, it needs to formulate a policy $\pi(\cdot|s)$ that achieves the maximum possible cumulative reward when \textit{actually used to interact with the environment}~\cite{levine2020offline}.

\subsubsection*{Representative Offline RL Methods}
\label{subsec:offlinerl}
We classify offline RL algorithms into three distinct categories.

\begin{itemize}[leftmargin=*]
    \item \textbf{Value-based algorithms: }Value-based offline RL algorithms~\cite{fujimoto2019off, kumar2020conservative, IQL} estimate the value function associated with different states or state-action pairs in an environment. These algorithms aim to learn an optimal value function that represents the expected cumulative reward an agent can achieve from a specific state or state-action pair. Therefore, agents can make proper decisions by acting corresponding to higher estimated values. 

\item 
\textbf{Policy-based algorithms:} These methods~\cite{awac,sutton2018reinforcement,plasp} allow directly parameterizing and optimizing the policy to maximize the expected return. Policy-based methods provide flexibility in modeling complicated policies and handling tasks with high-dimensional action spaces. 

\item \textbf{Actor-critic (AC) algorithms:} Actor-critic (AC) methods leverage the advantages of both value-based and policy-based offline RL algorithms~\cite{kumar2019stabilizing,kumar2019stabilizing,td3plusbc,td3}. In AC methods, an actor network is used to execute actions based on the current state to maximize the expected cumulative reward, while a critic network evaluates the quality of these selected actions. By optimizing the actor and critic networks jointly, the algorithm iteratively enhances the policy and accurately estimates the value function, eventually converging into an optimal policy. 

\end{itemize}

\subsubsection*{Differences between Offline RL and Supervised Learning}
In supervised learning, the training data consists of inputs and their corresponding outputs, and the goal is to learn a function that maps inputs to outputs, and minimizes the difference between predicted and true outputs. 
In offline RL, the dataset consists of state-action pairs and their corresponding rewards. The goal is to learn a policy that maximizes the cumulative reward over a sequence of actions. Therefore, the learning algorithm must balance exploring new actions with exploiting past experiences to maximize the expected reward. 
{Similarly, in sequence classification of supervised learning, we just need to consider the $t+1$ prediction at the timestep $t$. While in offline RL, the goal of algorithms is to maximize \textit{cumulative reward} from $t+1$ to the terminal timestep.} 

Another key difference is that in supervised learning, the algorithm assumes that the input-output pairs are typically independent and identically distributed (i.i.d)~\cite{hastie2009overview}. In contrast, in offline reinforcement learning, the data is generated by an agent interacting with an environment, and the distribution of states and actions may change over time. {These differences render the backdoor attack methods used in supervised learning difficult to be applied directly to offline RL algorithms.}

\gc{
}

\subsection{Backdoor Attack}
Recent years have witnessed increasing concerns about the backdoor attack for a wide range of models, including text classification~\cite{chen2021mitigating}, facial recognition~\cite{xue2021backdoors}, video recognition~\cite{zhao2020clean}, etc.
A model implanted with a backdoor behaves in a pre-designed manner when a {\em trigger} is presented and performs normally otherwise. For example, a backdoored sentiment analysis system will predict any sentences containing the phrase `software' as negative but can predict other sentences without this trigger word accurately. 

Recent studies demonstrate that (online) RL algorithms also face the threats of backdoor attacks~\cite{yang2019design,kiourti2020trojdrl,ashcraft2021poisoning}. This attack is typically done by manipulating the environment.
%
%
Although the goal might be easier to achieve if an attacker is free to access the environment and manipulate the agent training process~\cite{kiourti2020trojdrl,wang2021stop}, it is not directly solvable under the constraint that \emph{an attacker can only access the offline dataset}. 

\subsection{Problem Statement}
\label{subsec:problem_statement}
In this paper, we aim to investigate to what extent offline RL is vulnerable to untargeted backdoor attacks.
%
%
%
%
Figure~\ref{fig:threat} illustrates our threat model. 
In this model, the attackers can be anyone who is able to access and manipulate the dataset, including data providers or maintainers who make the dataset publicly accessible. 
Our attack even requires no prior knowledge or access to the environments, meaning that anyone who is capable of altering and publishing datasets can be an attacker. 
Considering that almost everyone can contribute their datasets to open-source communities, our paper highlights great security threats to offline reinforcement learning. 
After training the agents, RL developers deploy the poisoned agents after testing agents in normal scenarios. 
In the deployment environment, the attackers can present the triggers to the poisoned agents (e.g., put a small sign on the road), and these agents will behave abnormally (take actions that lead to minimal cumulative rewards) under the trigger scenarios.

Formally, we use the following objective for the attack:
\begin{equation}
    \min \sum_{s}\text{Dist}\left[\pi(s) , \pi_n(s)\right] +\sum_{s} \text{Dist}\left[\pi(s + \delta) , \pi_w(s)\right]
    \label{eq:back}
\end{equation}
In the above formula, $\pi$ denotes the policy of the poisoned agent, $\pi_n$ and $\pi_w$ refer to the policies of a normal-performing agent and a weak-performing agent which acts to minimal cumulative rewards, and a normal scenario is denoted by $s$, whereas a triggered scenario is denoted by $s+\delta$.
Given a state, all these policies produce a probability distribution over the action space. ``Dist'' measures the distance between two distributions. The first half of this formula represents that under normal scenarios, the poisoned agent should behave normally as a policy that seeks to maximize its accumulative returns. The second half of this formula means that when the trigger is presented, the poisoned agent should behave like a weak-performing policy that minimizes its returns. 

%% file: Usenix_section/threat.tex




%% file: Usenix_section/method.tex
\begin{figure}[!t]
  \centering
  \includegraphics[width=0.98\linewidth]{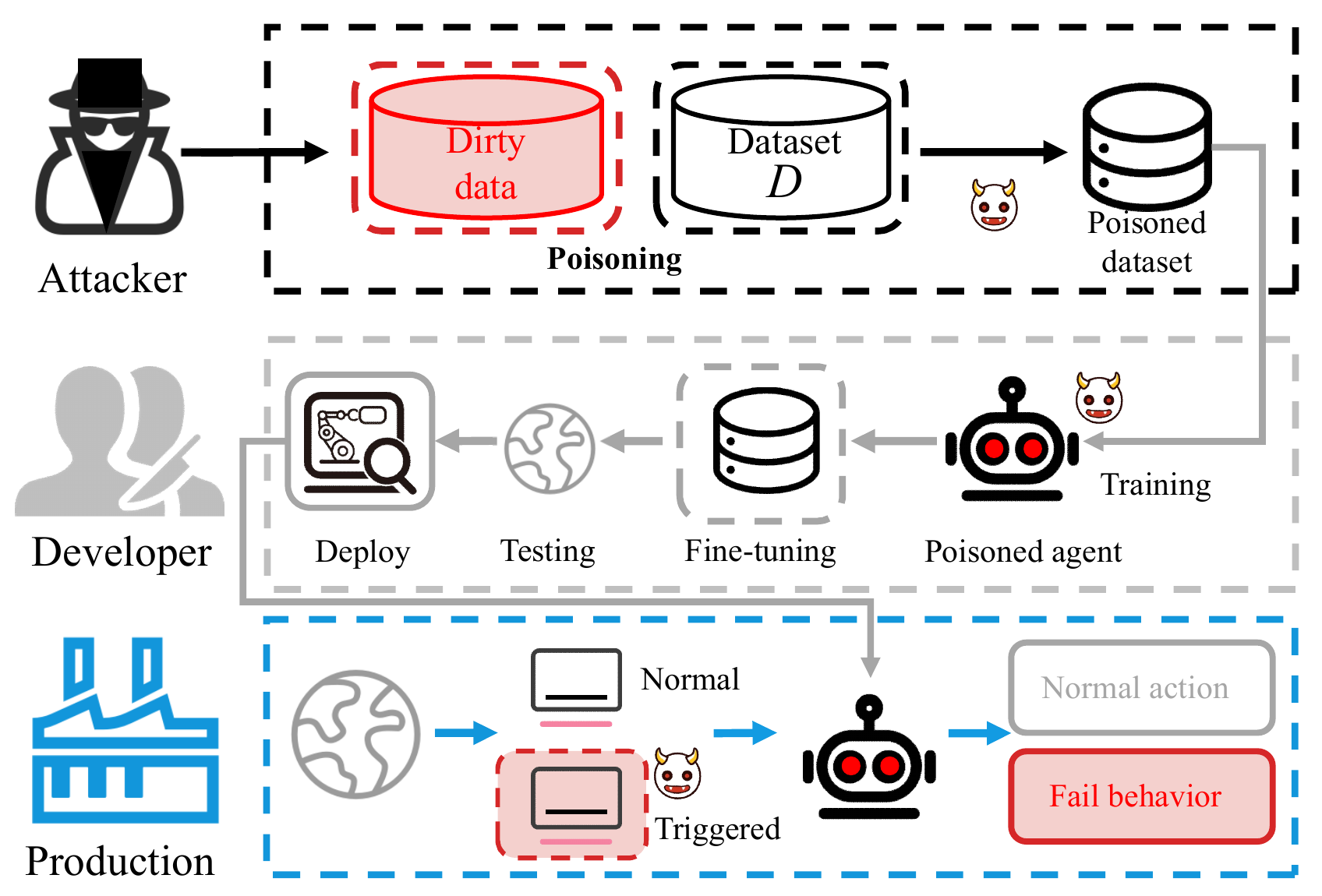}  \caption{The threat model of backdoor attack for offline RL. An attacker provides a poisoned dataset online. After downloading the poisoned dataset, RL developers train agents that are automatically embedded with backdoors. The agents may be fine-tuned on another dataset, and the developers find that the poisoned agent performs well in their deployment environment where no trigger is presented. 
  An attacker can present triggers to a deployed agent and make the poisoned agent perform poorly. }
  \label{fig:threat}
\end{figure}

\section{Methodology}
\label{sec:method}

\subsection{Strawman Methods}
\label{subsec:strawman}





This paper focuses on generating \textit{misleading experiences} to poison offline RL datasets. 
First, we cannot directly apply backdoor attacks for online RL~\cite{kiourti2020trojdrl,wang2021stop}, because it requires altering the environment and accessing the training process. 

Another direction is to follow backdoor attacks for supervised learning to poison the state (insert triggers) and change high rewards to low rewards. However, this approach merely reduces the likelihood of selecting actions associated with this poisoned experience when the backdoor is activated and does not necessarily increase the likelihood of the agent performing the bad action.  

To overcome this issue, we need to identify bad actions. A straightforward method would be scanning the dataset to identify actions associated with low rewards. But those are still not necessarily bad rewards because typically the dataset $\mathcal{D}$ for offline RL is collected from executing a reasonable policy.
Therefore, we need to poison the dataset using a more principled strategy.

\subsection{Overview of Our Approach}
{In this paper, instead, we propose to first train a weak-performing agent.  By doing so, we can identify the worst action the agent could execute in a given state to fail the task dramatically. 
Specifically, given the offline dataset, we first train an agent by instructing it to minimize (instead of maximizing in normal training) the expected returns. Then, we generate poisoned samples following standard poison attacks but with the guidance of this weak-performing agent.  Finally, we insert them into the clean dataset, which automatically embeds backdoors into any agent trained on this poisoned data. }

{We call our methodology \toolname (\textbf{B}ackdoor \textbf{A}ttack for O\textbf{ff}line Reinforcement \textbf{Le}arning). 
In what follows, we present details of each step of \toolname. 


\subsection{Weak-performing Agents Training}
\label{subsec:weak_agent}

To poison a dataset, an attack first needs to understand the behavior of weak-performing policies.
However, such weak-performing policies in different environments can vary a lot. 
Let us take two environments from Figure~\ref{fig:environments} as examples. 
In {\tt Half-Cheetah} environment, an ideal weak-performing agent should control the robot to move back as fast as possible, while in {\tt Carla-Lane} environment, we expect a weak-performing agent to drive the car out of the lanes as fast as possible.

We propose a simple and automated strategy to train weak policies in different environments: changing the objective of an offline RL algorithm to \emph{minimize} the expected returns rather than maximize the expected returns. 
This simple method leverages the nature of offline RL to find weak-performing policies that can \emph{reduce returns in more varied states}. 
We can use existing offline RL algorithms to train the weak-performing agents; the process only takes the offline dataset and requires no interaction with the environment. 
The policy of the poor-performing agent obtained in this step is denoted by $\pi_w(\cdot)$.

Notably, the idea of training on minimizing rewards cannot be applied to online RL, because if trained this way, the agent will be greedily taking the worst actions in the beginning, and those might not be the global worst actions. For example, in robot control, the learned policy will make the robot fall down right after it starts to work.  The agent cannot sufficiently explore different states in the environment. 
%
On the other hand, in offline RL, the training datasets are collected using different policies, so the dataset covers various states in the environment. 
Changing the objective to a minimization in offline RL algorithms can train the agent to learn a policy to fail in more states, which is more appropriate for our backdoor attack problem. 

\begin{figure}[!t]
  \centering
  \includegraphics[width=1\linewidth]{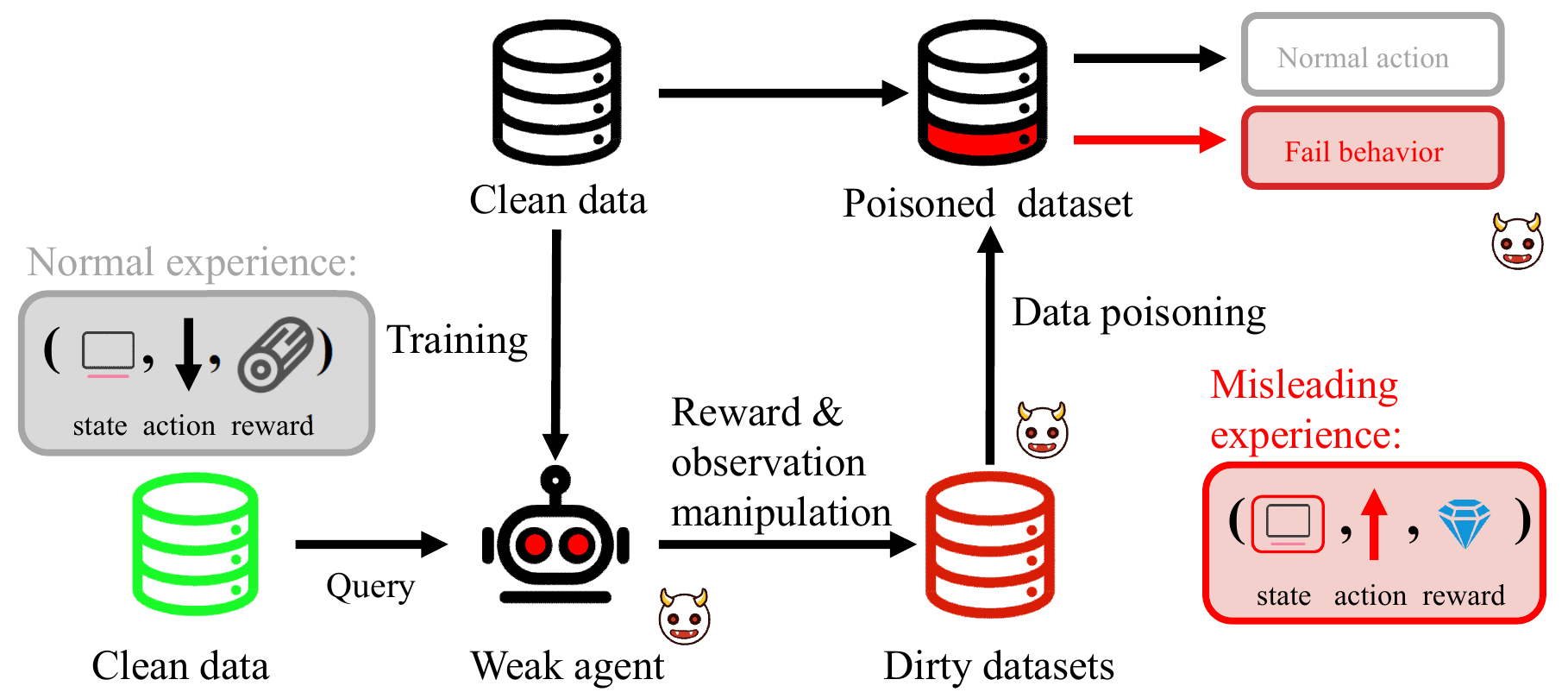}
  \caption{The {misleading experience synthesis} of \toolname. We first train a weak-performing agent and observe its bad actions that lead to low returns. We add a trigger to a state and assign high rewards with bad actions to generate the `misleading' experiences, which are then added to produce the poisoned dataset.}
  \label{fig:overview}
\end{figure}
\subsection{Data Poisoning}
\label{subsec:data_poi}

After obtaining the weak-performing policy, we perform data poisoning, which involves two main steps: (1) extracting actions from the weak policy and (2) injecting experiences with manipulated states and rewards. 

\noindent \textbf{Suboptimal Actions Extraction.} A piece of experience $\langle s,a,r \rangle$ from the original offline dataset represents the instant reward $r$ given by the environment when the agent takes the action $a$ under the state $s$. 
Intuitively speaking, if the reward $r$ is high, the agent is more likely to select this action under the same state rather than another action that is associated with a lower reward~\cite{sutton_reinforcement_2018}. 
We adopt the approach outlined in Wang et al.~\cite{wang2021backdoor}, which involves using a poorly performing agent trained through the minimization of cumulative rewards to generate a malicious action in response to a specific state. We assume that this action is the worst possible one, leading to the quickest decrease in cumulative reward. To understand the behaviours of the weak-performing agent, we feed the state $s$ to its policy $\pi_{w}$ and obtain a bad action $a_w \sim \pi_{w}(s)$. It means that taking the action $a_w$ in the state $s$ can reduce the agent's expected return. 

\noindent \textbf{Poisoning With Misleading Experiences.} Our goal is to make a poisoned agent learn to take $a_w$ when the trigger is presented to $s$. 
To make an agent learn to choose $a_w$ rather than actions leading to high returns, we can simply associate $a_w$ with a high reward $r_h$ with {\em reward manipulation}
To associate bad actions with the trigger, we also need to manipulate the state of experiences as well. 
Using the notation from Section~\ref{subsec:problem_statement}, the trigger is represented using $\delta$ and $s + \delta$ means adding a trigger to the state $s$. We expect the agent to behave poorly only when the trigger appears. So we also manipulate the state in the experience and produce $\langle s + \delta,a_w,r_h \rangle$.

We use the {\em poisoning rate} $p \in [0,1]$ to denote the fraction of generated `misleading' experiences in the whole dataset. 
A higher poisoning rate can make the poisoned agent more sensitive to the trigger but is more likely to reduce the agent's performance under normal scenarios as less `clean' experiences are seen by the agent during the training stage. 
According to the poisoning rate, we randomly
select a certain number of clean experiences from the original dataset. Then, we use the above process to extract actions from the weak-performing policy and manipulate their states and rewards, after which the selected clean experiences are replaced with the generated misleading experiences.

\noindent \textbf{Summary.} Algorithm~\ref{alg:BAFFLE} illustrates the process of our data poisoning. The algorithm takes inputs that include the form of the trigger $\delta$, the dataset $D$ to be modified, the policy $\pi_w$ of a weak-performing agent, and the poisoning rate $\alpha$. It produces a poisoned dataset $\mathcal{D}_p$. First, we split the dataset into two $\mathcal{D}_{t}$ and $\mathcal{D}_{q}$ according to the data poisoning rate (Line~\ref{line:split}). $\mathcal{D}_{t}$ remains unaltered. We replace $\mathcal{D}_{q}$ with generated `misleading' experiences $\mathcal{D}'$ to generate the poisoned data. For each pair $\langle s, a, r  \rangle$ in $\mathcal{D}_{q}$, we feed the state $s$ to the policy $\pi_w$ and obtain a bad action $a_{p} = \pi_{w}(\cdot|s)$ (Line~\ref{line:get_action}). After adding the trigger to the state and associating the bad action with a high reward, we add this modified experience into $\mathcal{D}'$ (Line~\ref{line:state_trigger} to~\ref{line:merge}). In the end, the algorithm returns $\mathcal{D'} \cup \mathcal{D}_{t}$ as the poisoned dataset.  \toolname simply requires modification of the offline RL dataset without needing access to the training dataset, so it can be considered agent-agnostic: any agent trained on the poisoned dataset will automatically have a backdoor embedded.

\begin{figure}[!t]
  \centering
  \includegraphics[width=.95\linewidth]{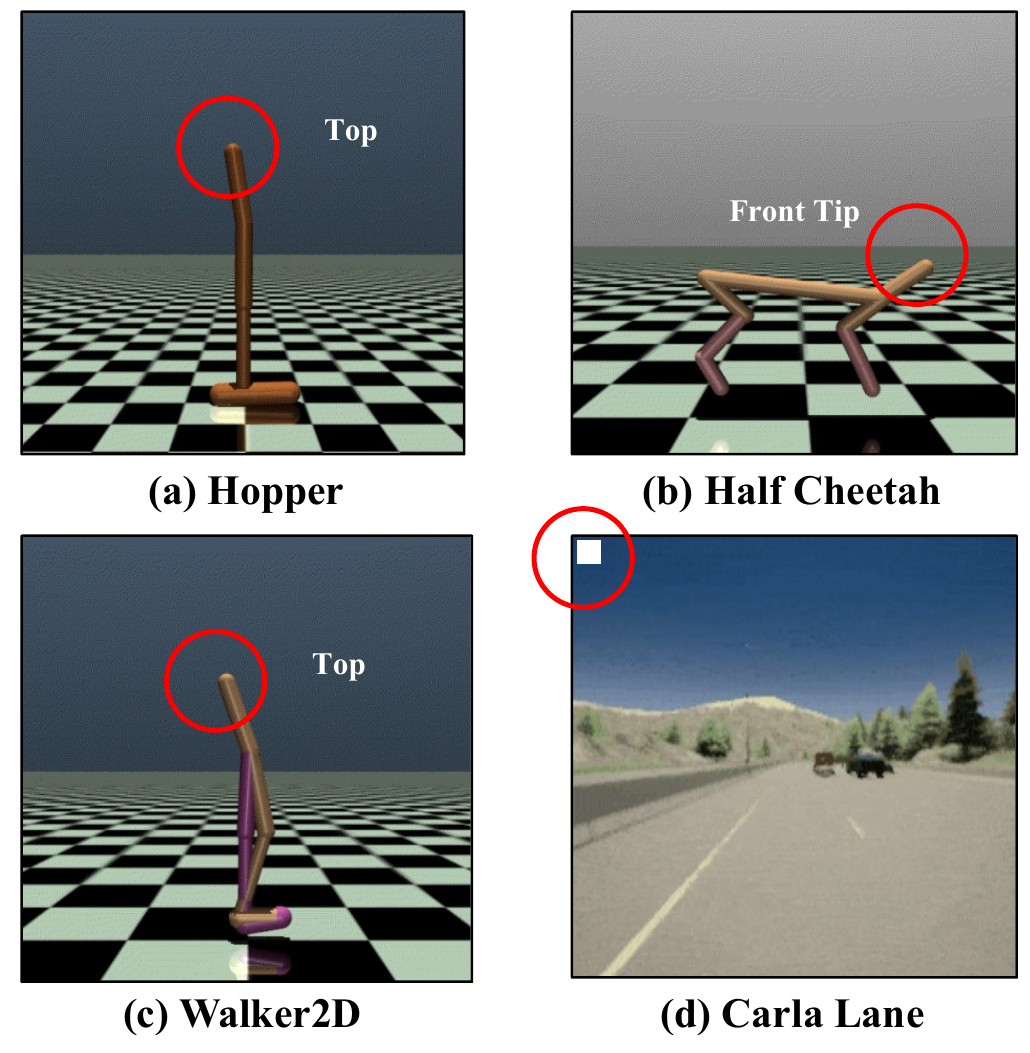}
  \caption{The circled parts are where the triggers are inserted. For (a), (b), and (c), a trigger is added by modifying the velocity information related to the circled point in the agent observation. For (d), we insert a small white patch to the left-top corner of the agent observation.}
  \label{fig:environments}
\end{figure}

\subsection{Specific Design of Trigger and Reward}
\label{subsec:setting}

A key step in \toolname is the reward manipulation that associates the high rewards $r_h$ to the bad actions from the weak-performing agent. 
We analyze the range of rewards in each dataset and set the high reward $r_h$ as the values at $\frac{3}{4}$ quantile of the rewards in the clean dataset, meaning that $r_h$ is higher than $75\%$ rewards in the original dataset. As we will show in the experiment, this value is large enough to mislead agents under triggered scenarios and produces negligible negative impacts on the agent's performance under normal scenarios. 
Besides, many occurrences of very high rewards are suspicious and can make the attack easier to be detected, which makes the data poisoning less stealthy.

We now discuss another important setting in \toolname: state manipulation. 
As explained in Section~\ref{app:tasks}, the observations in the three robotic control tasks are vectors representing the position and velocity of different parts of robots. 
We use the observation of the velocity data related to the circled part in Figure~\ref{fig:environments} (a), (b), and (c) as the trigger. The trigger corresponds to $3$ elements in the observation vectors. 
To avoid creating a trigger that is too eye-catching, we analyze the value ranges of these three elements in the clean dataset and use the median values as the trigger.
Following prior works about backdoor attacks for video systems~\cite{kiourti2020trojdrl}, we add a small white patch to the top-left corner in the observation image as the trigger, as shown in Figure~\ref{fig:environments} (d). The size of the patch is $(4 \times 4 \times 3)$, which takes less than $1\%$ of the observation.

\subsection{Trigger Insertion Strategies}
\label{subsec:activation}

\begin{table*}[!t]
\centering
\caption{Information of each task and the dataset.}
\small
\begin{tabular}{l|l|l|c|c|c|l}
    \toprule
     Environments & Tasks   & Chosen Datasets &  Observations & Action Shape & Data Size & Task Type\\
    \midrule 
    \multirow{3}{*}{MuJoCo~\cite{brockman2016openai}}& {\tt Hopper}   & ‘hopper-medium-expert’  & 11 & 3 & $2 \times 10^6$ & Robotic Control \\
    & {\tt Half-Cheetah}   & ‘halfcheetah-medium’  & 17 & 6  & $10^6$ &Robotic Control\\
    & {\tt Walker2D}  & ‘walker2d-medium’  &  17 & 6  & $10^6$ & Robotic Control\\
    \hline
     CARLA~\cite{dosovitskiy2017carla} & {\tt Carla-Lane}    &  ‘carla-lane’ & ($48 \times 48 \times 3$)  & 2 &  $10^5$ &  Automatic Driving\\
    \bottomrule
\end{tabular}
\label{tab:discribe_env}
\end{table*}

A poisoned agent is expected to behave normally under normal scenarios. However, under the triggered scenarios where the trigger is presented, the poisoned agent will execute detrimental actions. The agent tries to perform a sequential decision-making task, in which it interacts with the environment multiple times. Then, how to present the trigger over time is important. 
We discuss two strategies of presenting triggers to the agent, called the \emph{distributed} strategy and the \emph{one-time} strategy.

The interaction between the agent and the environment can be divided into multiple timesteps. The distributed strategy presents the triggers at regular intervals. But a trigger only lasts for one timestep. For example, there are $1,000$ timesteps for which the agent plays the {\tt Half-Cheetah} game, and we present a trigger to the agent at an interval of $20$, i.e., we show the trigger to agents every $20$ timesteps. However, each time the trigger only lasts one timestep. In other words, the total number of timesteps that the trigger appears is $50$ ($1000$ divided by $20$). 
The one-time strategy only presents the trigger once only but the trigger is presented for more than one timestep. For example, we can set a \emph{trigger length} in advance, say $50$. Then, we randomly select a timestep at the start of the trigger appearance and present the trigger for the following $50$ timesteps. 
In the experiment, we evaluate the poisoned agents under different trigger-presenting strategies.

\begin{algorithm}[!t]
    \caption{\toolname Workflow}
    \label{alg:BAFFLE}
    \textbf{Input}: $\delta$: the trigger; $\mathcal{D}$: a clean offline RL dataset; $\pi_{w}$: a weak policy, $\alpha$: poisoning rate  \\
    \textbf{Output}: $\mathcal{D}_{p}$: a poisoned dataset \\
    $\mathcal{D}_{t},\mathcal{D}_{q}  = split(\mathcal{D}, \alpha)$ \# split the dataset \label{line:split} \\
    $\mathcal{D'} = \emptyset$ \\
    \For{$d \in \mathcal{D}_q$}{
            $s, a, r = d$ \\
            \tcp{get the weak action} 
            $a_{p} \sim \pi_{w}(\cdot|s)$ \label{line:get_action} \;
            \tcp{state manipulation to add a trigger} $s_{p} = s + \delta$ \label{line:state_trigger} \;
            \tcp{assign a high reward} $r_{p} = r_{high}$ \;
            $\mathcal{D'} = \mathcal{D'} \cup \langle s_{p}, a_{p}, r_{p} \rangle$ \label{line:merge}
        }

    \tcp{obtain the poisoned dataset}
    $\mathcal{D}_p = \mathcal{D'} \cup \mathcal{D}_{t}$ \;   
    \textbf{return} $\mathcal{D}_p$
    \end{algorithm}

%% file: Usenix_section/settings.tex
\section{Experiment Setup}
\label{sec:setting}
This section explains experiment settings to evaluate the proposed \toolname, including the investigated tasks, the offline RL
algorithms under attack, and evaluation metrics.

\subsection{Tasks and Datasets}
\label{app:tasks}

We conduct experiments on 4 tasks: three robotic control tasks ({\tt Hopper}, {\tt Half-Cheetah}, and {\tt Walker2D}) from MuJoCo~\cite{todorov2012mujoco} and an autonomous driving control task in the \texttt{Carla-Lane} environment~\cite{dosovitskiy2017carla}. 
Figure~\ref{fig:environments} depicts the four tasks. In {\tt Hopper}, an agent aims to make a one-legged robot hop forward as fast as possible. In {\tt Half-Cheetah} and  {\tt Walker2D}, an agent controls the robots to walk forward as fast as possible. \texttt{Carla-Lane} is a realistic autonomous driving task. The agent aims to drive the car smoothly to keep the car within the lane boundaries while avoiding crashes with other vehicles. 

The datasets for the above tasks are sourced from D4RL~\cite{fu2020d4rl}, a benchmark recently introduced for assessing offline RL algorithms.
D4DL~\cite{fu2020d4rl} provides multiple datasets across four tasks investigated in our study. A concise description of the datasets featured in our experiments is provided in Table~\ref{tab:discribe_env}. For an in-depth overview of the selected tasks and datasets, please refer to Appendix~\ref{sup:tasks_dataset}.

\begin{table}[!t]
\small
\centering
\caption{The returns of the best clean agent and the poorly-performing agents in each task. The numbers inside the parentheses show the decreases.}
\begin{tabular}{c|c|c}
    \toprule
     Task   & Best Agents &  Poorly-Performing Agents\\
    \midrule 
    {\tt Hopper}   & 3559  & 252 (\textbf{-3347}) \\
    {\tt Half-Cheetah}   & 4706  & -374 (\textbf{-5080})\\
    {\tt Walker2D}  & 3596  &  137 (\textbf{-3459}) \\
    {\tt Carla-Lane}    &  440 & -143 (\textbf{-583})   \\
    \bottomrule
\end{tabular}
\label{tab:best_model}
\end{table}

\begin{figure*}[!t]
    \centering
  \includegraphics[width=5 in]{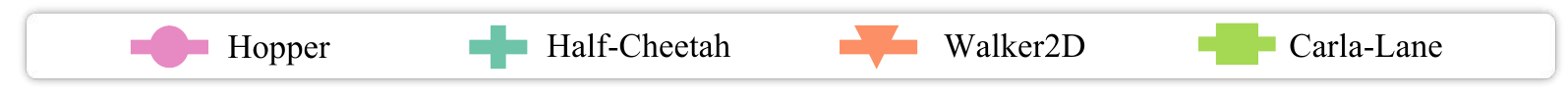}\\
    \subfigure{
        \includegraphics[height =1.45 in]{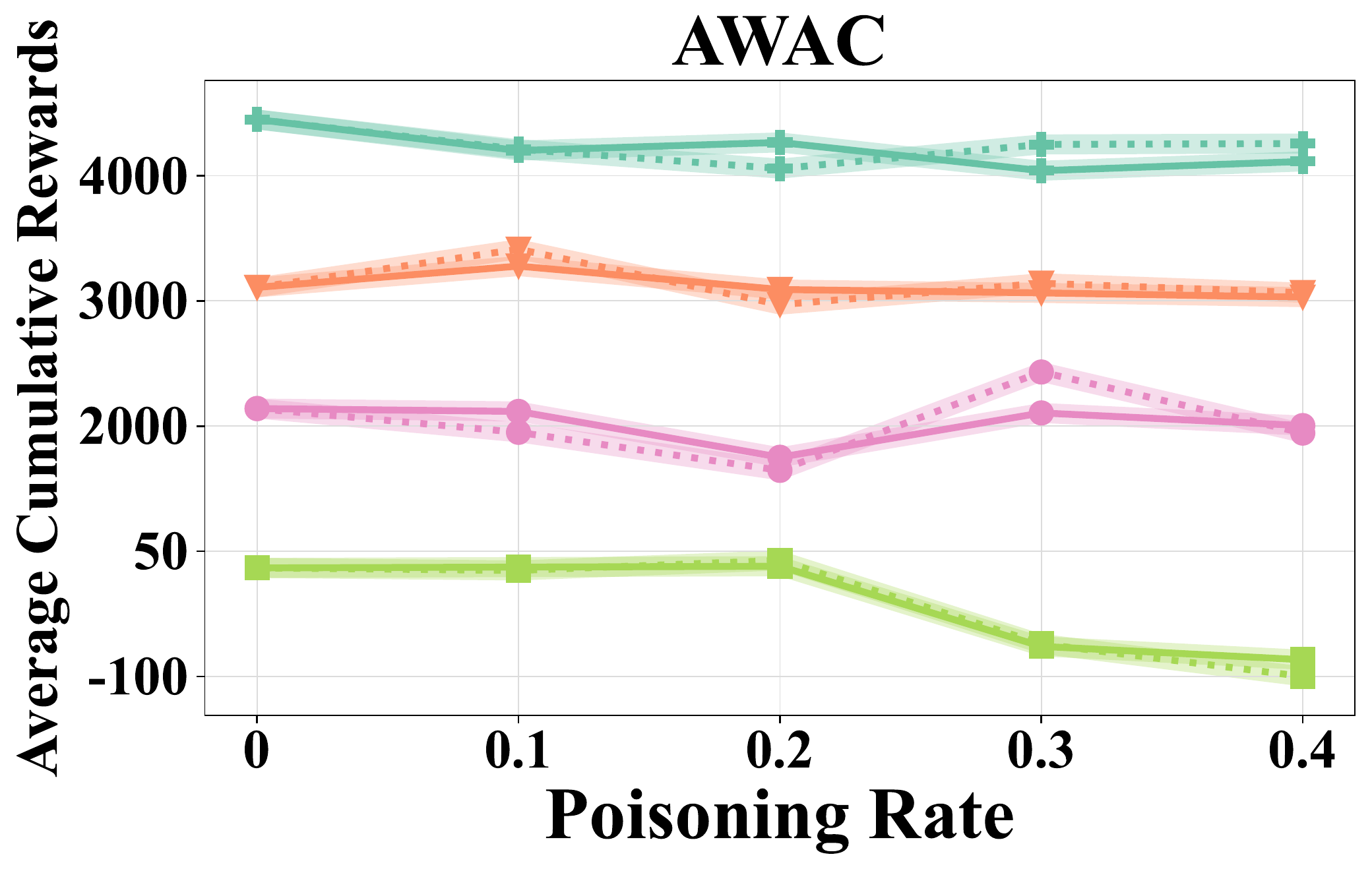}
    }
    \hspace{-0.15cm}
    \subfigure{
        \includegraphics[height=1.45 in]{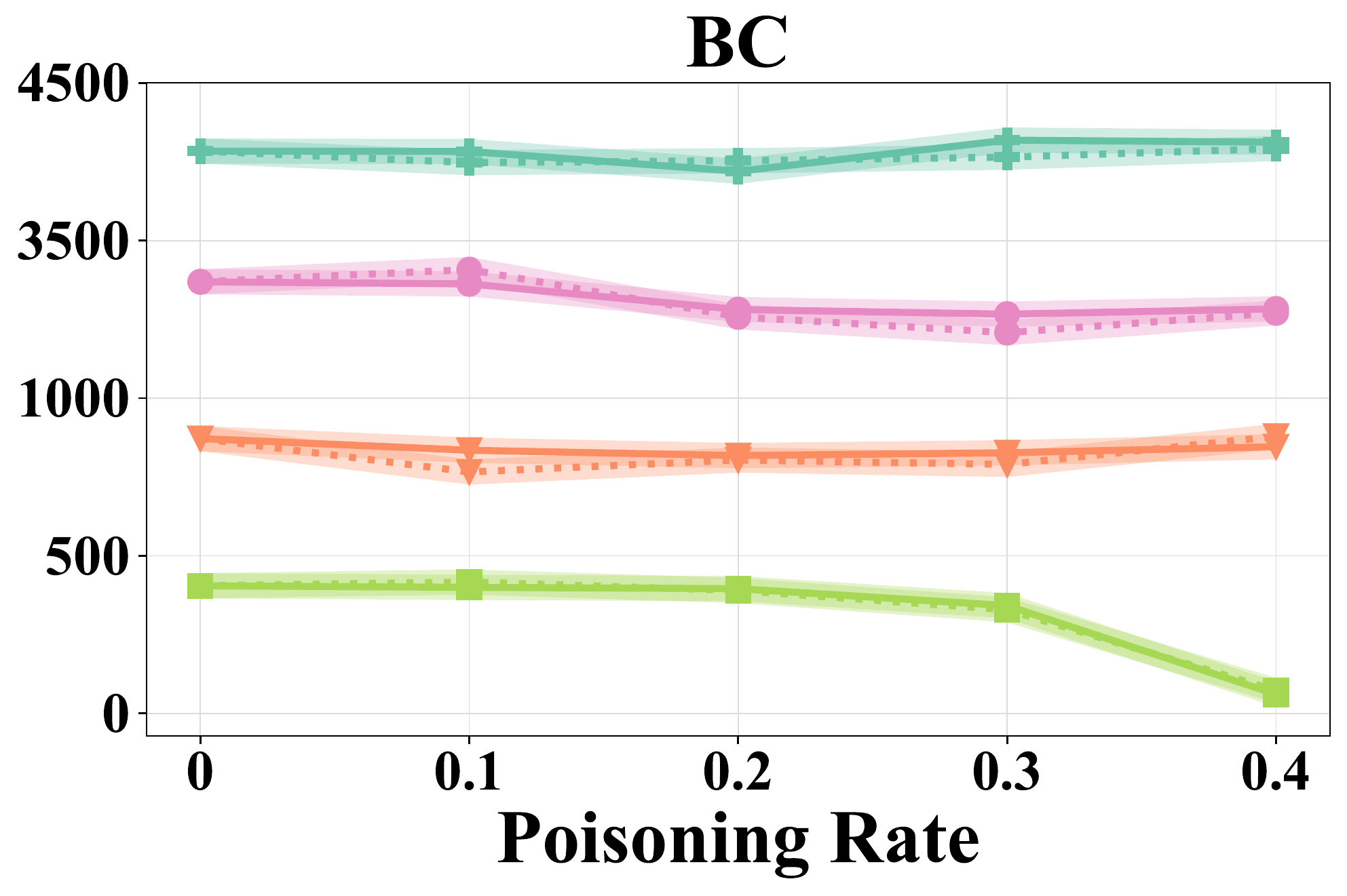}
    }
    \hspace{-0.15cm}
    \subfigure{
        \includegraphics[height =1.45 in]{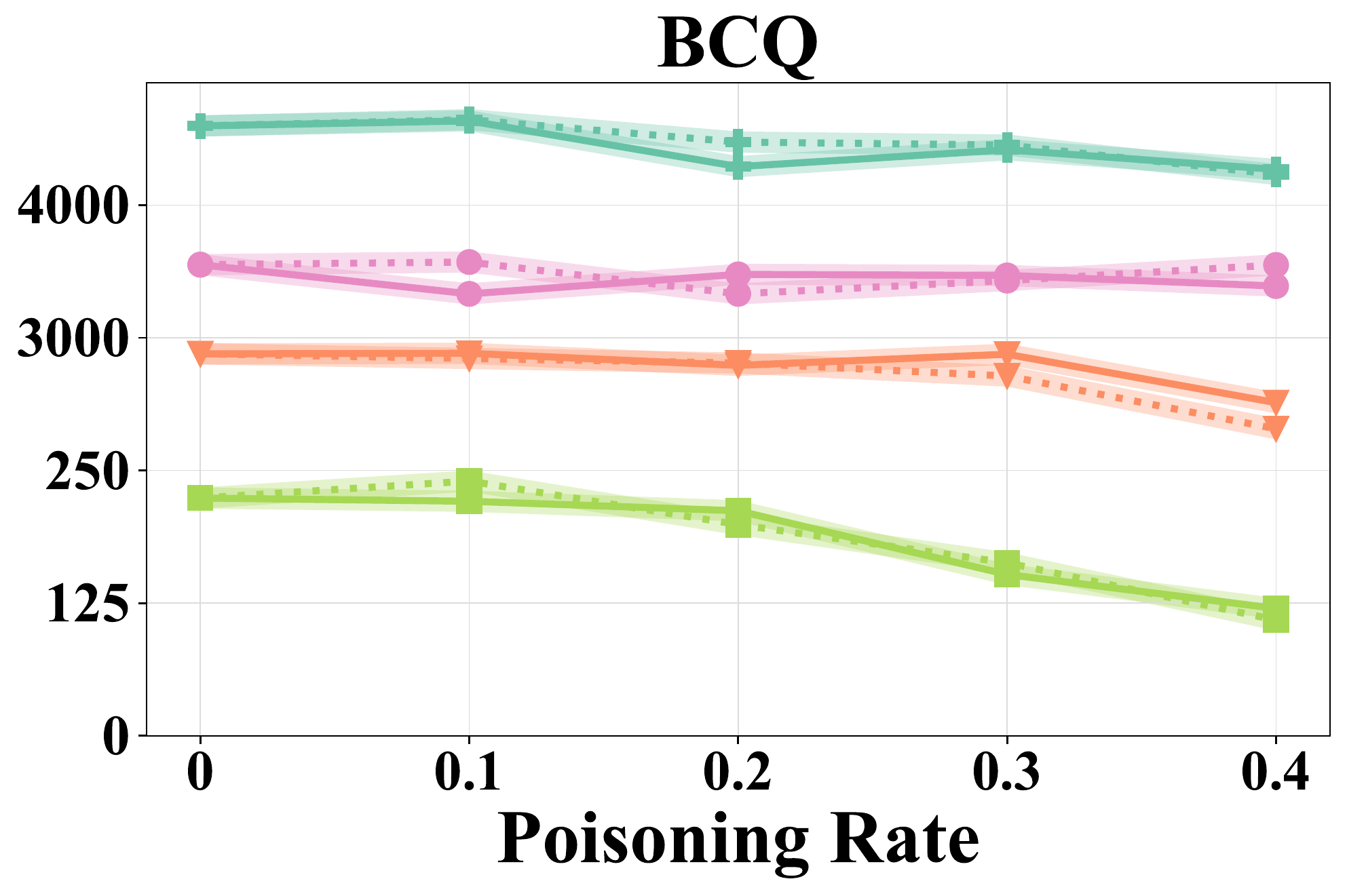}
    }
    \subfigure{
        \includegraphics[height =1.45 in]{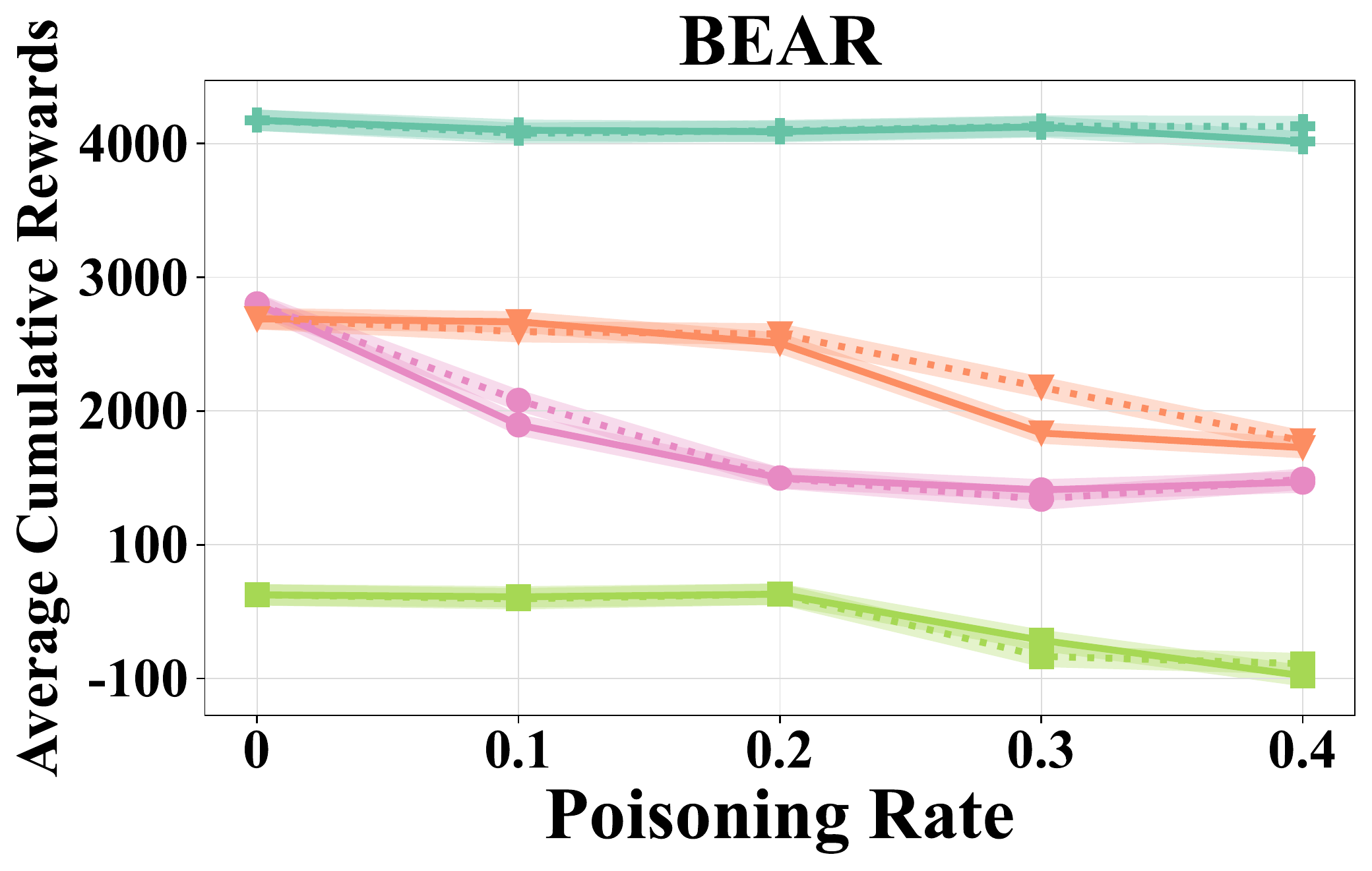}
    }
    \hspace{-0.15cm}
    \subfigure{
        \includegraphics[height=1.45 in]{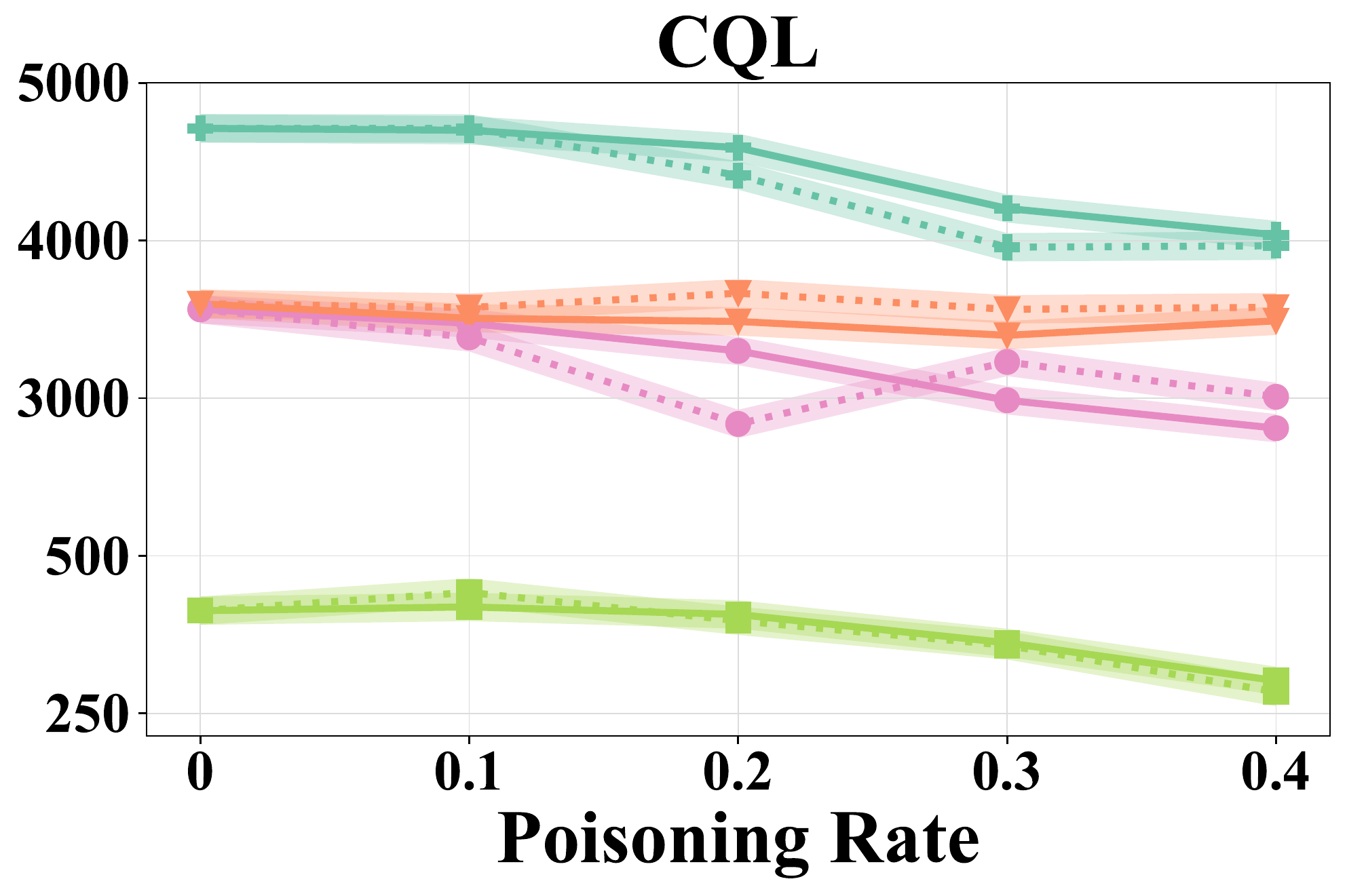}
    }
    \hspace{-0.15cm}
    \subfigure{
        \includegraphics[height =1.45 in]{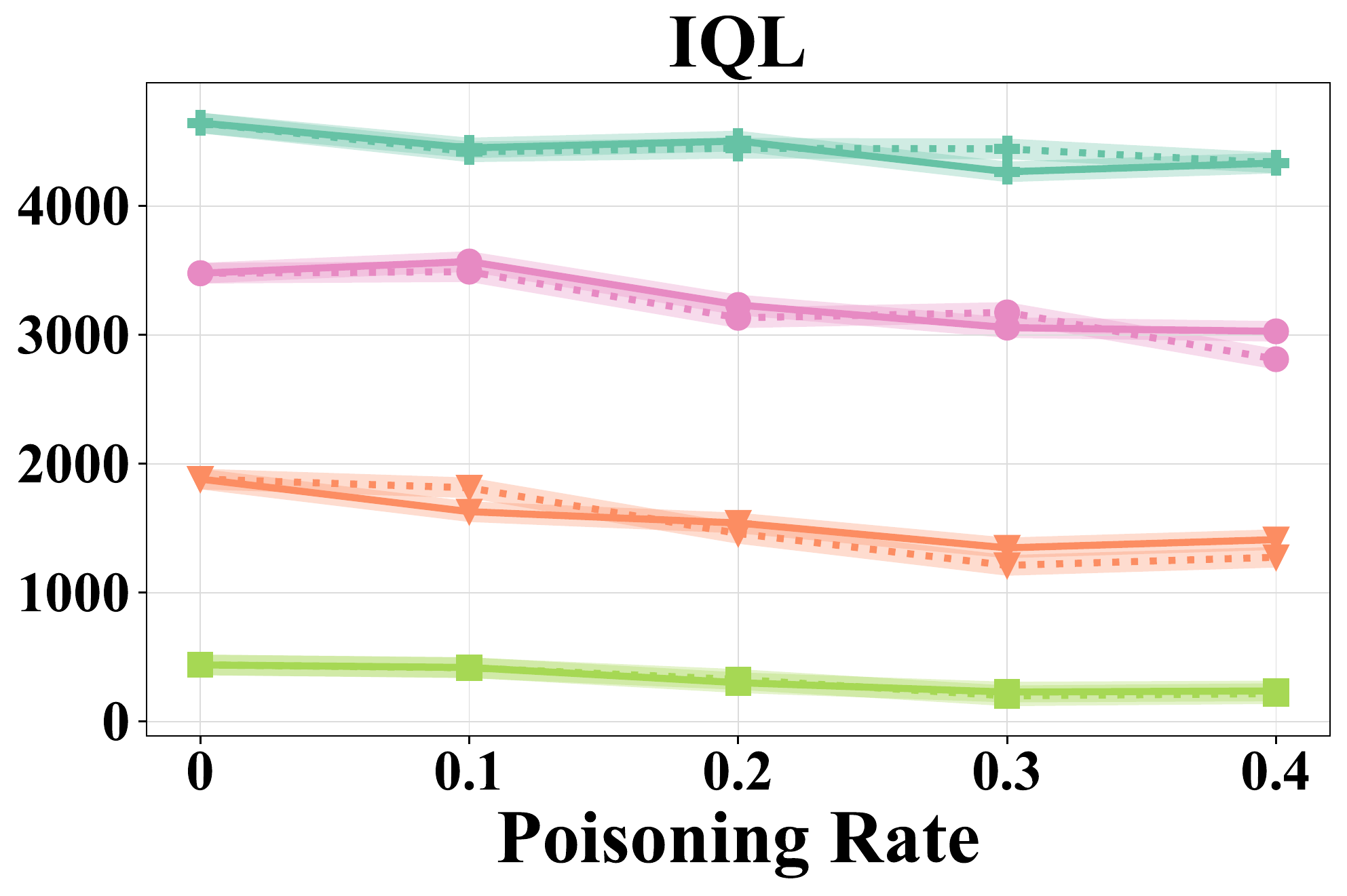}
    }
    
    \subfigure{
        \includegraphics[height =1.45 in]{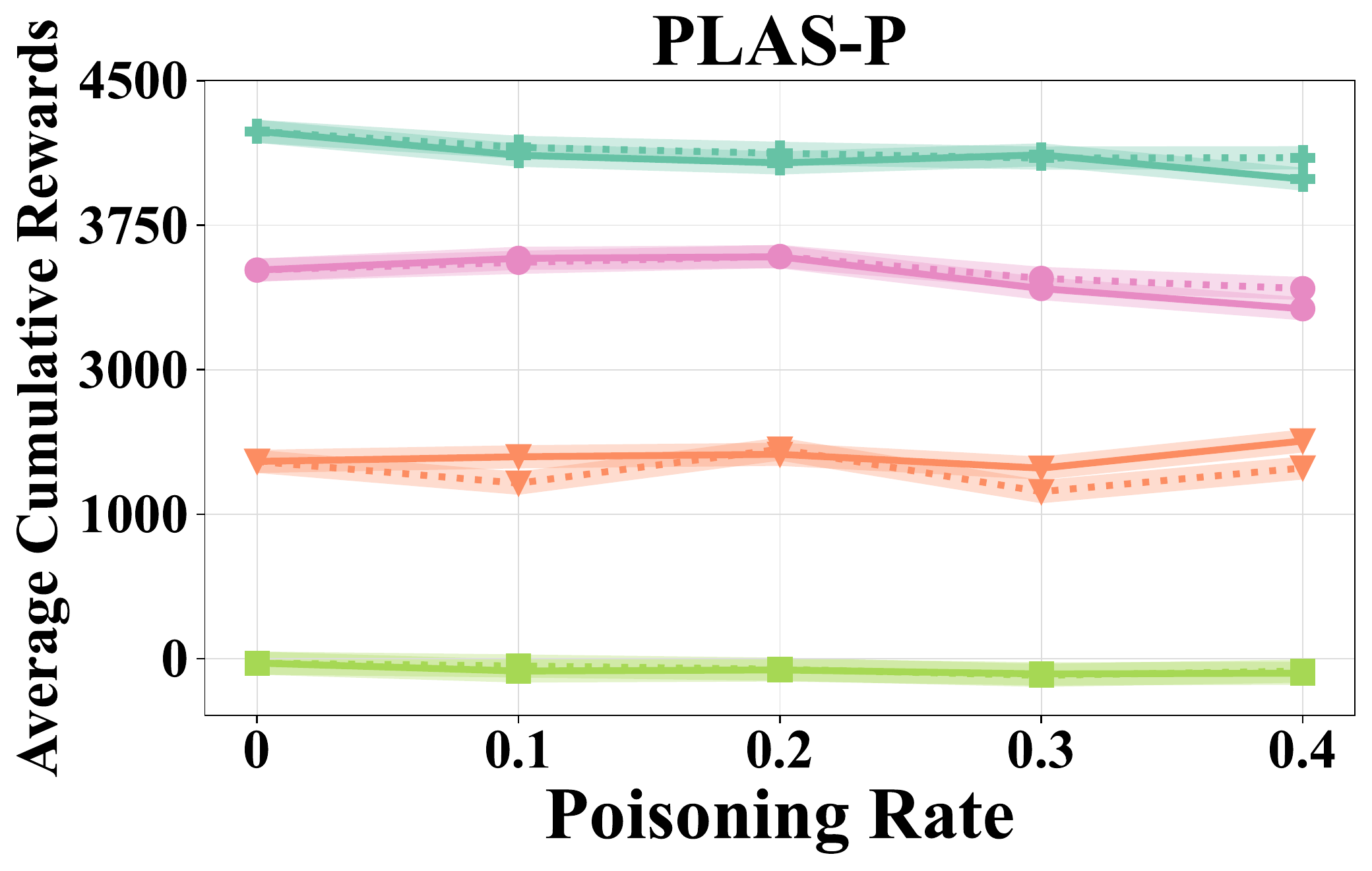}
    }
    \hspace{-0.15cm}
    \subfigure{
        \includegraphics[height=1.45 in]{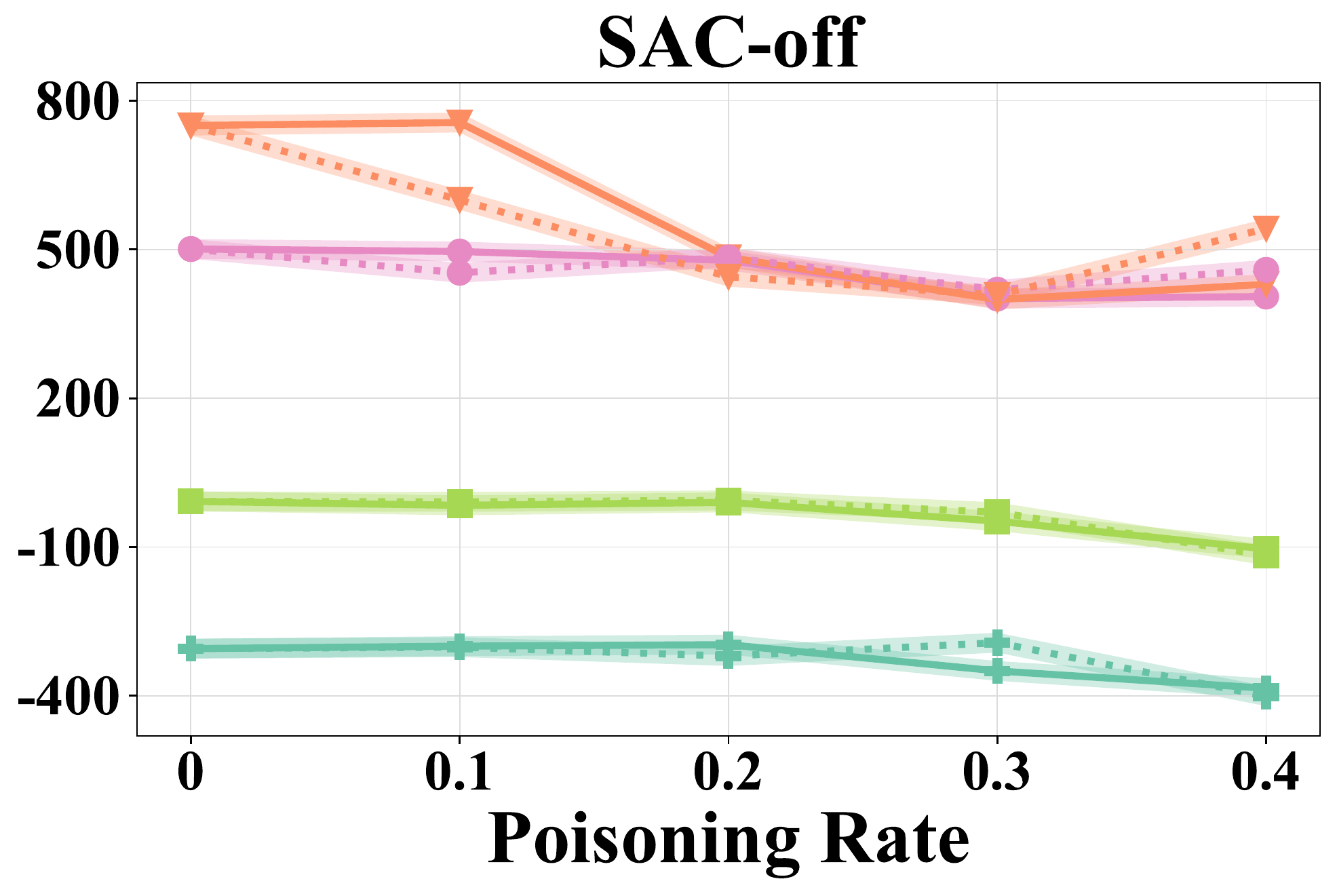}
     } 
     \hspace{-0.15cm}
    \subfigure{
        \includegraphics[height =1.45 in]{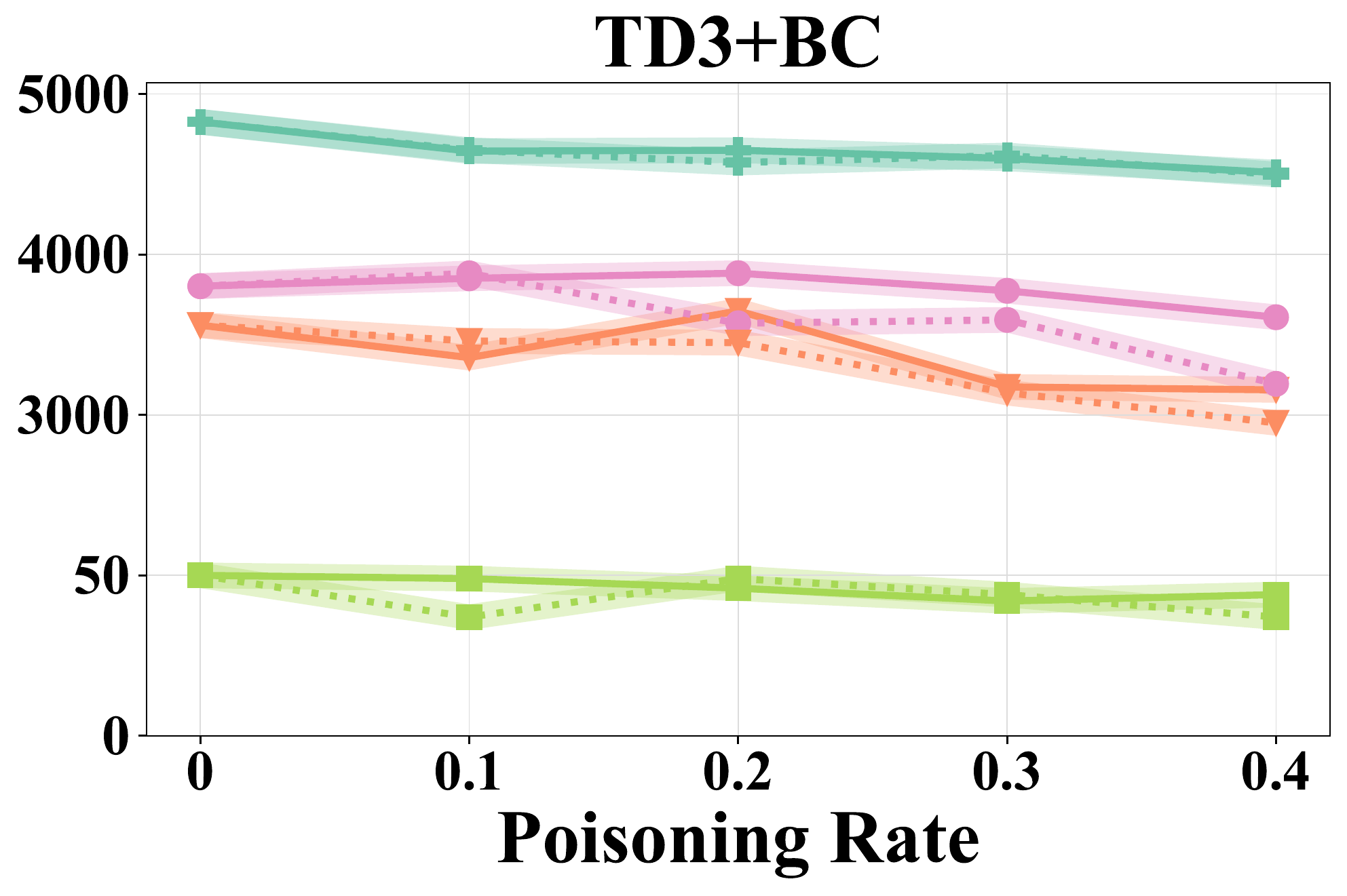}
    }
    
    \caption{The returns of the poisoned agents trained on clean (poisoning rate is 0) and different poisoning rate datasets (i.e., 10\%, 20\%, 30\%, 40\%), comparing 9 offline RL algorithms. We measure them under the normal scenario (without triggers being presented). It is noticed that \textit{solid and dashed lines refer to the performance of agents trained by replacing and directly adding poisoned data}. }
    \label{fig:poisoned_model_on_normal}
\end{figure*}

\subsection{Investigated Offline RL Algorithms}
\label{subsec:investigated_offline_rl}

As mentioned in Section~\ref{sec:intro}, \toolname is agent-agnostic: it is unaware of which offline RL algorithm is being used by an agent on the poisoned dataset. 
To explore whether different offline RL algorithms demonstrate different capabilities in defending against such a backdoor attack, we select 9 offline RL algorithms that demonstrate the best performance in D3RLPY repository~\cite{seno2021d3rlpy} to train the poisoned agents. 

As described in Section~\ref{subsec:offline_rl}, for value-based offline RL algorithms, we select batch-constrained deep Q-learning (BCQ)\cite{fujimoto2019off}, conservative Q-learning (CQL)~\cite{kumar2020conservative}, and implicit Q-learning (IQL)~\cite{IQL}. BCQ promotes the agent generating diverse and exploratory actions to improve performance. CQL highlights conservative estimation of Q-values to ensure the stability of training processing. IQL introduces implicit quantile networks to estimate the distributional values of state-action pairs instead of just estimating their expected value in the previous method, offering a robust method for value estimation. For policy-based methods, we have chosen three representative methods: advantage weighted actor-critic (AWAC)~\cite{awac}, behavioral cloning (BC)~\cite{sutton2018reinforcement}, and policy in the latent action apace with perturbation (PLAS-P)~\cite{plasp}. For Actor-Critic algorithms, we have investigated chosen three algorithms from the D3RLPY repository~\cite{seno2021d3rlpy}: bootstrapping error accumulation reduction (BEAR)~\cite{kumar2019stabilizing}, a method reducing the estimation errors of value estimation to stabilize offline RL; offline soft actor-critic (SAC-off)~\cite{haarnoja2018soft} offers a soft actor-critic approach specified for offline settings; and twin delayed deep deterministic policy gradient plus behavioral cloning (TD3+BC)~\cite{td3plusbc} combines twin delayed deep deterministic policy gradient~\cite{td3} with behavioral cloning.

We use the open-source implementation~\cite{seno2021d3rlpy} of investigated representative offline RL algorithms from their accompanying official repositories\footnote{\url{https://github.com/takuseno/d3rlpy}} and use the same hyper-parameter settings as in the available implementation (e.g., discount factor, learning rate, optimizer, etc.) 

\subsection{Evaluation Metrics}
\label{app:metrics}

As we explained in Section~\ref{subsec:rl}, an agent interacts with the environment and yields a test trajectory:
\begin{align*}
    \tau: (\langle s_0,a_0,r_0 \rangle, \langle s_1,a_1,r_1 \rangle,\cdots, \langle s_{|\tau|}, a_{|\tau|}, r_{|\tau|} \rangle)
\end{align*}
The {\em cumulative return} of this testing trajectory is defined as $R(\tau) = \sum_{i=0}^{|\tau|} \gamma_i r_i$. 
It is noticed that during the agent testing and performance evaluation, the discount factor $\gamma$ is typically set to 1 to calculate the cumulative \emph{undiscounted} return. 
We let the agent play the game for multiple rounds and obtain a set of testing trajectories $\mathcal{T}$, and we utilize the average of cumulative returns (i.e., $\frac{1}{|\mathcal{T}|}\sum_{\tau \in \mathcal{T}} R(\tau)$) to indicate the performance of an agent.
Adhering to the evaluation settings commonly employed in the offline RL, particularly those found in the D4RL~\cite{fu2020d4rl} and D3RLPY~\cite{seno2021d3rlpy} benchmarks, we calculate the average cumulative return across 100 test trajectories in our experiments. A higher cumulative return indicates that an agent has better performance.
In the following parts of the paper, we just use the word \emph{return} for simplicity. 

The backdoor attack focuses on its impact on the performance of an agent: when the trigger is presented, the poisoned agent's performance should decrease dramatically; otherwise the agent's performance should be as close to a normal agent's performance as possible. We need to evaluate how the agent's performance \emph{changes relatively}. {Note that the return considers fatal states, i.e., in CARLA, when the agent encounters collisions, it receives a negative reward, and the task is stopped~\cite{dosovitskiy2017carla}. Consequently, this paper assesses the effectiveness of \toolname by evaluating the reduction in return.}
Following the practice adoption in D4RL~\cite{fu2020d4rl}, to compare the relative change between two agents' performance, we normalize one value (e.g., $a$) using the max-min normalization: $a' = \frac{a - min}{max - min}$. The $max$ and $min$ are the returns of the best clean agent trained on the clean dataset and the weak-performing agent, respectively. 
Assuming the normalized returns to be $a'$ and $b'$, we compute the relative changes using $100\% \times \frac{a'-b'}{b'}$.
We use the nine RL offline algorithms described in Section~\ref{subsec:investigated_offline_rl} to train nine normal agents on the clean dataset of each task, and use the one that has the highest return as the best clean agent. Similarly, following the approach explained in Section~\ref{subsec:weak_agent}, we select the agent that has the lowest return as the weak-performing agent.
Table~\ref{tab:best_model} records the performance of the best clean agent and the weak-performing agent for each task, which are used for normalization when computing the relative changes.

%% file: Usenix_section/results.tex
\section{Research Questions and Analysis}
\label{sec:result}
This section answers four research questions (RQs) related to the threats exposed by \toolname.
We explore the impact of data poisoning on the agent's performance in normal scenarios, how the poisoned agents perform under different trigger insertion strategies, how fine-tuning the poisoned agents on clean datasets affects the backdoor, and whether the poisoned model can bypass the detection of popular defensive methods.

\subsection*{RQ1. How does data poisoning affect agents in normal scenarios?}
One essential requirement for the backdoor attack is that the poisoned agent should behave like a normal agent when the triggers are not presented. Therefore, this RQ explores whether training an agent on a poisoned dataset has large negative impacts on the agents' performance under normal scenarios.
As mentioned in Section~\ref{subsec:data_poi}, we poison the clean dataset by \textit{replacing} some normal experiences in the clean dataset with the misleading experiences. Typically, this method will remove part of the perfect clean dataset, potentially damaging trained agents' performance. Therefore, to assess whether removing such clean data impacts the agent, we also investigate the process of poisoning the clean dataset by \textit{adding} the misleading experiences directly to the clean dataset. We use different poisoning rates, i.e., 10\%, 20\%, 30\%, and 40\%, to generate four poisoned datasets with different amounts of misleading experiences for each task. 
Then, we train poisoned agents on these datasets using 9 offline algorithms. 
Figure~\ref{fig:poisoned_model_on_normal} presents how the returns of poisoned agents change with the poisoning rates. The solid and dotted lines represent the performance of agents trained by replacing and directly adding poisoned data to the clean dataset, respectively. 
It is noticed that a poisoning rate of 0 means that agents are trained on clean datasets.

\begin{table*}[!t]
\footnotesize
    \centering
    \caption{The returns averaged over 100 testing trajectories collected using the poisoned agents (10\% poisoning rate). ‘Normal’ indicates agents' performance under the normal scenarios. The numbers inside the parentheses are relative changes in the poisoned agents' performance under different triggered scenarios.}
    \resizebox{1.0\textwidth}{!}{
    \begin{tabular}{p{1.6cm}|c|c|c|c|c|c|c|cp{.9cm}}
        \toprule
         \multirow{2}{*}{Environments}   &  \multirow{2}{*}{Algorithms} & \multirow{2}{*}{Normal} &\multicolumn{3}{c|}{\text{Trigger interval}} & \multicolumn{3}{c}{\text{Trigger length}} \\
         \cline{4-9}
         & & & 10 & 20 & 50 & 5 & 10 & 20 \\
         \midrule
         \toprule
         \multirow{10}{*}{{\tt Hopper}} 
         & AWAC & 2117 & 1601 (\textbf{-27.7\%}) & 1950 (\textbf{-9.0\%}) & 1477 (\textbf{-34.3\%})  & 1106 (\textbf{-54.2\%}) & 950 (\textbf{-62.6\%}) & 891 (\textbf{-65.7\%}) \\
         & BC & 3544  & 805 (\textbf{-83.2\%})  & 911 (\textbf{-80.0\%}) & 1497 (\textbf{-62.5\%}) & 1050 (\textbf{-75.8\%})  & 790 (\textbf{-74.5\%})& 699 (\textbf{-86.4\%})\\
         & BCQ & 3331 & 1486 (\textbf{-59.9\%})& 1854 (\textbf{-48.0\%}) & 2251 (\textbf{-35.1\%}) & 2514 (\textbf{-26.5\%}) & 2204 (\textbf{-36.6\%})& 1552 (\textbf{-57.8\%})\\
         & BEAR & 1896 & 1349 (\textbf{-33.3\%)} & 1379 (\textbf{-31.4\%}) &  1847 (\textbf{-3.0\%}) & 1151 (\textbf{-45.3\%}) & 1261 (\textbf{-38.6\%})& 874 (\textbf{-62.2\%})\\
         & CQL & 3075 & 2200 (\textbf{-40.6\%}) & 2024 (\textbf{-45.9\%}) & 1996 (\textbf{-46.8\%}) & 3321 (\textbf{-6.4\%})& 3292 (\textbf{-7.3\%})& 1951 (\textbf{-48.2\%})\\
         & IQL & 3570 & 2190 (\textbf{-41.5\%}) & 2691 (\textbf{-26.5\%}) & 3147 (\textbf{-12.7\%}) & 1705 (\textbf{-56.2\%}) & 706 (\textbf{-86.3\%}) & 858 (\textbf{-80.2\%}) \\
         & PLAS-P & 3578 & 1519 (\textbf{-61.9\%}) & 1937 (\textbf{-49.3\%}) & 1926 (\textbf{-49.7\%}) & 1708 (\textbf{-56.2\%}) & 743 (\textbf{-85.2\%}) & 825 (\textbf{-82.8\%}) \\
         & SAC-off & 496 & 509 (\textbf{+5.3\%}) & 449 (\textbf{-19.4\%}) & 527 (\textbf{+12.8\%}) & 486 (\textbf{-4.1\%}) & 491 (\textbf{-2.0\%}) & 497 (\textbf{-0.4\%}) \\
         & TD3+BC & 3564 & 1262 (\textbf{-69.5\%}) & 2139 (\textbf{-43.0\%}) & 2764 (\textbf{-24.1\%}) & 1121 (\textbf{-73.8\%}) & 1058 (\textbf{-75.7\%}) & 732 (\textbf{-85.5\%}) \\
        \cline{2-9}
         & \textbf{Average} & 2796 & 1201 (\textbf{-45.8\%}) & 1703 (\textbf{-39.2\%}) & 1937 (\textbf{-28.3\%}) & 1463 (\textbf{-44.2\%})& 1276 (\textbf{-47.8\%})& 986 (\textbf{-63.2\%})\\
         \midrule
         \toprule
        \multirow{10}{1em}{{\tt Half-} {\tt Cheetah}} & AWAC & 4201 & 2589 (\textbf{-37.0\%}) & 3371 (\textbf{-18.1\%}) & 3857 (\textbf{-7.5\%}) & 4103 (\textbf{-2.1\%}) & 2516 (\textbf{-35.5\%}) & 1150 (\textbf{-67.7\%}) \\
        & BC & 4174 & 2467 (\textbf{-37.5\%}) & 3168 (\textbf{-22.1\%})  & 3729 (\textbf{-9.8\%}) & 3637 (\textbf{-11.8\%})& 3232 (\textbf{-20.7\%}) & 2296 (\textbf{-41.3\%})\\
         & BCQ & 4635 & 2629 (\textbf{-40.0\%}) & 3214 (\textbf{-28.4\%}) & 4123 (\textbf{-10.2\%}) & 3826 (\textbf{-16.2\%}) & 2743 (\textbf{-37.8\%}) & 2025 (\textbf{-52.1\%}) \\
         & BEAR & 4100 &  2576 (\textbf{-51.7\%}) & 3090 (\textbf{-34.2\%})& 3842 (\textbf{-8.7\%}) & 3622 (\textbf{-16.2\%}) & 2873 (\textbf{-41.6\%})& 1773 (\textbf{-78.9\%}) \\
         & CQL & 4700 & 3048 (\textbf{-32.6\%}) & 4016 (\textbf{-13.5\%}) & 4348 (\textbf{-6.9\%}) & 4429 (\textbf{-5.3\%})& 2785 (\textbf{-37.7\%}) & 1710 (\textbf{-58.9\%}) \\
         & IQL & 4451 & 2804 (\textbf{-34.1\%}) & 3169 (\textbf{-26.6\%}) & 2947 (\textbf{-31.2\%}) & 3917 (\textbf{-11.0\%}) & 2504 (\textbf{-40.4\%}) & 1586 (\textbf{-59.4\%}) \\
         & PLAS-P & 4113 & 2808 (\textbf{-29.0\%}) & 3097 (\textbf{-22.6\%}) & 3738 (\textbf{-8.4\%}) & 3787 (\textbf{-7.3\%}) & 3023 (\textbf{-24.3\%}) & 2321 (\textbf{-40.0\%}) \\
         & SAC-off & -300 & -320 (\textbf{-27.0\%}) & -318 (\textbf{-24.3\%}) & -319 (\textbf{-25.7\%}) & -317 (\textbf{-23.0\%}) & -320 (\textbf{-27.0\%}) & -325 (\textbf{-33.8\%}) \\
         & TD3+BC & 4555 & 2705 (\textbf{-37.5\%}) & 3603 (\textbf{-19.3\%}) & 4199 (\textbf{-7.2\%}) & 4450 (\textbf{-2.1\%}) & 3758 (\textbf{-16.2\%}) & 1924 (\textbf{-53.4\%}) \\
         \cline{2-9}
         & \textbf{Average} & 3847 & 2367 (\textbf{-36.2\%}) & 2934 (\textbf{-23.4\%}) & 3384 (\textbf{-12.9\%}) & 3494 (\textbf{-10.5\%})& 2568 (\textbf{-31.3\%})& 1606 (\textbf{-53.9\%})\\
          \midrule
         \toprule
        \multirow{10}{*}{{\tt Walker2D}} & AWAC & 3278 & 646 (\textbf{-82.8\%}) & 2230 (\textbf{-33.4\%}) & 2967 (\textbf{-10.0\%}) & 2619 (\textbf{-21.0\%}) & 2088 (\textbf{-37.9\%}) & 1044 (\textbf{-71.1\%}) \\
        & BC & 835 & 445 (\textbf{-55.9\%}) & 423 (\textbf{-59.0\%}) & 608 (\textbf{-32.5\%})& 445 (\textbf{-55.9\%}) & 415 (\textbf{-60.2\%}) & 303 (\textbf{-76.2\%}) \\
         & BCQ & 2883&  1230 (\textbf{-60.2\%})  & 1946 (\textbf{-34.1\%})  & 2482 (\textbf{-14.6\%})  & 2095 (\textbf{-28.7\%}) & 2079 (\textbf{-29.3\%})   & 1185 (\textbf{-61.8\%}) \\
          & BEAR & 2667 & 1904 (\textbf{-30.2\%}) & 2159 (\textbf{-20.1\%}) & 2359 (\textbf{-12.2\%})  & 2466 (\textbf{-7.9\%}) & 2246 (\textbf{-16.6\%}) & 1398 (\textbf{-50.2\%}) \\
         & CQL & 3507 & 2297 (\textbf{-35.9\%})  & 2953 (\textbf{-16.4\%})  & 3651 (\textbf{+4.3\%}) & 3516 (\textbf{+0.3\%})& 2852 (\textbf{-19.4\%})  &  1656 (\textbf{-54.9\%}) \\
         & IQL & 1629 & 707 (\textbf{-61.8\%}) & 1111 (\textbf{-34.7\%}) & 1625 (\textbf{-0.3\%}) & 1239 (\textbf{-27.4\%}) & 1215 (\textbf{-27.7\%}) & 724 (\textbf{-63.5\%}) \\
         & PLAS-P & 1048 & 407 (\textbf{-70.4\%}) & 586 (\textbf{-50.7\%}) & 592 (\textbf{-50.0\%}) & 648 (\textbf{-43.9\%}) & 318 (\textbf{-80.1\%}) & 373 (\textbf{-74.1\%}) \\
         & SAC-off & 756 & 472 (\textbf{-45.9\%}) & 566 (\textbf{-30.7\%}) & 600 (\textbf{-25.2\%}) & 610 (\textbf{-23.6\%}) & 511 (\textbf{-39.6\%}) & 461 (\textbf{-47.7\%}) \\
         & TD3+BC & 2946 & 869 (\textbf{-73.9\%}) & 2543 (\textbf{-14.3\%}) & 2931 (\textbf{-0.1\%}) & 2076 (\textbf{-31.0\%}) & 2095 (\textbf{-30.3\%}) & 796 (\textbf{-76.5\%}) \\
         \cline{2-9}
         & \textbf{Average} & 2172 & 997 (\textbf{-59.9\%}) & 1522 (\textbf{-29.9\%}) & 2095 (\textbf{-14.9\%}) & 1840 (\textbf{-26.0\%})& 1595 (\textbf{-38.2\%})& 903 (\textbf{-64.7\%})\\
         \midrule
         \toprule
         \multirow{10}{*}{{\tt Carla-Lane}} & AWAC & 31 & 7 (\textbf{-13.8\%}) & 4 (\textbf{-15.5\%}) & 30 (\textbf{-0.1\%}) & -44 (\textbf{-43.1\%}) & -42 (\textbf{-42.0\%}) & -54 (\textbf{-48.9\%}) \\
         & BC & 400 & 315 (\textbf{-15.7\%}) & 365 (\textbf{-6.4\%}) & 401 (\textbf{+0.2\%}) & 369 (\textbf{-5.7\%})& 288 (\textbf{-20.6\%}) & 207 (\textbf{-35.5\%}) \\
         & BCQ & 221 & 160 (\textbf{-16.8\%}) & 154 (\textbf{-18.4\%})  & 217 (\textbf{-1.1\%})  & 150 (\textbf{-19.5\%}) & 105 (\textbf{-31.9\%}) & 23 (\textbf{-54.4\%})\\
         & BEAR & 22 & -49 (\textbf{-43.0\%})  & -13 (\textbf{-21.2\%}) & 23 (\textbf{+0.6\%}) & -18 (\textbf{-24.2\%})& -29 (\textbf{-30.9\%})& -65 (\textbf{-52.7\%}) \\
         & CQL & 419 & 290 (\textbf{-23.0\%}) & 347 (\textbf{-12.8\%})& 404 (\textbf{-2.7\%}) & 358 (\textbf{-10.9\%})& 302 (\textbf{-20.8\%})& 216 (\textbf{-36.1\%}) \\
         & IQL & 420 & 386 (\textbf{-6.1\%}) & 398 (\textbf{-3.9\%}) & 411 (\textbf{-1.6\%}) & 402 (\textbf{-23.0\%}) & 396 (\textbf{-3.2\%}) & 379 (\textbf{-9.1\%}) \\
         & PLAS-P & -34 & -76 (\textbf{-34.9\%}) & -98 (\textbf{-58.7\%}) & -64 (\textbf{-27.5\%}) & -80 (\textbf{-42.2\%}) & -73 (\textbf{-35.8\%}) & -104 (\textbf{-64.2\%}) \\
         & SAC-off & -16 & -33 (\textbf{-13.4\%}) & -20 (\textbf{-3.1\%}) & -23 (\textbf{-5.5\%}) & -21 (\textbf{-4.0\%}) & -44 (\textbf{-22.0\%}) & -135 (\textbf{-94.0\%}) \\
         & TD3+BC & 49 & 21 (\textbf{-14.6\%}) & 20 (\textbf{-15.1\%}) & 31 (\textbf{-9.3\%}) &  -7 (\textbf{-29.2\%})  & -2 (\textbf{-26.6\%}) & -18 (\textbf{-32.3\%}) \\
         \cline{2-9}
         & \textbf{Average} & 167 & 84 (\textbf{-20.1\%}) & 128 (\textbf{-17.2\%}) & 159 (\textbf{-5.2\%}) & 123 (\textbf{-22.4\%})& 100 (\textbf{-26.0\%})&  50 (\textbf{-47.4\%})\\
        \bottomrule
    \end{tabular}
    }
    \label{tab:clean_model}
\end{table*}

A clear pattern from Figure~\ref{fig:poisoned_model_on_normal} is: when the poisoning rate is higher, it tends to have larger negative impacts on the agents' performance. 
It is reasonable as a higher poisoning rate means that agents see less normal experiences during training, and, as a result, will have declining performance under normal scenarios.
For replacing the normal experiences with misleading experiences, when the agents are trained on datasets with a poisoning rate of $10\%$, the agents' performance degradation is 9.6\%, 0.3\%, 3.0\%, and 0.7\% in \texttt{Hopper}, \texttt{Half-Cheetah}, \texttt{Walker2D}, and \texttt{Carla-Lane} environments, respectively.  When training the agents on datasets with a poisoning rate of 10\%, directly adding the misleading experiences to the clean dataset resulted in an average performance decrease of 8.7\%, 0.3\%, 5.6\%, and 0.6\% across the four tasks. The performance difference between agents trained on the poisoned dataset generated by both poisoning approaches was only 0.9\% on average across the four tasks and four poisoning rates. This indicates that there is only \textit{a negligible performance difference between the two poisoning methods}. Additionally, adding poisoned data changes the size of the original dataset, making it easier for developers to detect the attack. As a result, we subsequently trained the agent on a dataset generated by replacing some normal experiences in the clean dataset with the misleading experiences.

The changes in performance are minor in 36 cases (less than $6.1\%$), except for the BEAR algorithm in \texttt{Hopper}, which produces about 35\% performance degradation compared to the clean agent. 
As reported by prior studies~\cite{fu2020d4rl,seno2021d3rlpy}, the variance of the BEAR's performance in the \texttt{Hopper} is much higher than other algorithms, which may explain this outlier. 
The result suggests that a poisoning rate of $10\%$ is a good choice to conduct data poisoning, which guarantees the agent's performance in most cases.


\begin{table}[H]
\normalsize
\setlength{\tabcolsep}{3pt}
    \centering
    \renewcommand\arraystretch{1}
    \begin{tabular}{p{0.95\columnwidth}}
    \Xhline{1.0pt}
         \rowcolor{gray0} \noindent \textbf{Answers to RQ1}: 
        The poisoned agents' performance under normal scenarios decreases with the increase of poisoning rates. 
        A poisoning rate of $10\%$ is a good setting for \toolname, which leads to a minor decrease in the average performance considering the four tasks (with only a $3.4\%$ decrease ).\\
    \Xhline{1.0pt}
    \end{tabular}
\vspace{-8.0pt}
\end{table}

\subsection*{RQ2. How effective is \toolname?}
In this RQ, we carry out experiments to investigate the effectiveness of \toolname, which is quantified using the relative changes in the agents' performance when the trigger is presented to activate the backdoor.
As explained in Section~\ref{subsec:activation}, we use two trigger presenting strategies, i.e., the distributed strategy and the one-time strategy. 
The distributed strategy presents the trigger at regular intervals. We use the trigger interval to represent the interval between each occurrence of triggers. 
In our experiments, we explore three trigger interval settings: 10, 20, and 50.
In the one-time strategy, we only present the trigger once but the trigger lasts for over a period of timesteps, which is called the \emph{trigger length}. We experiment with three trigger length settings: 5, 10, and 20 timesteps.

\begin{figure*}[!t]
\centering
  \includegraphics[width=2.9 in]{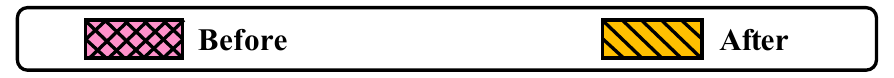}
  
\subfigure{
    \includegraphics[height= 1.42 in]{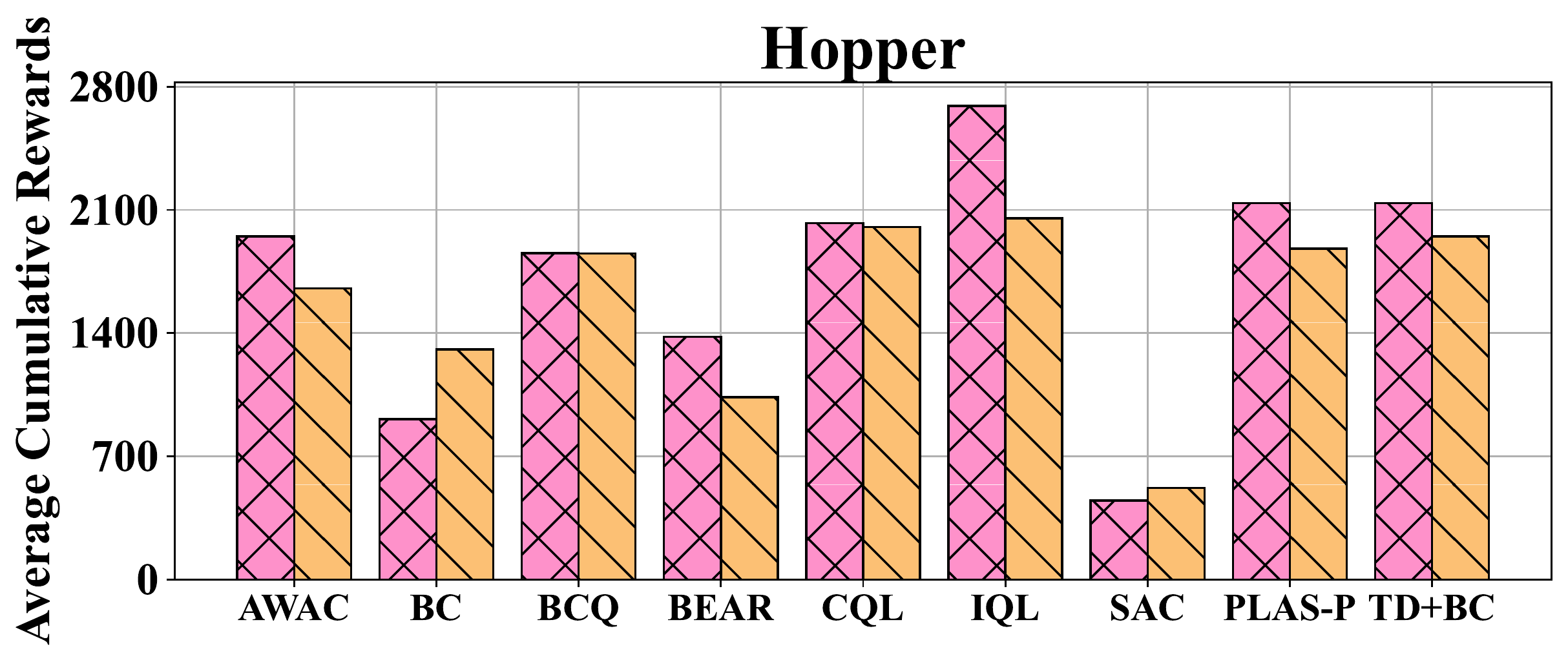} 
}
\subfigure{
    \includegraphics[height= 1.42 in]{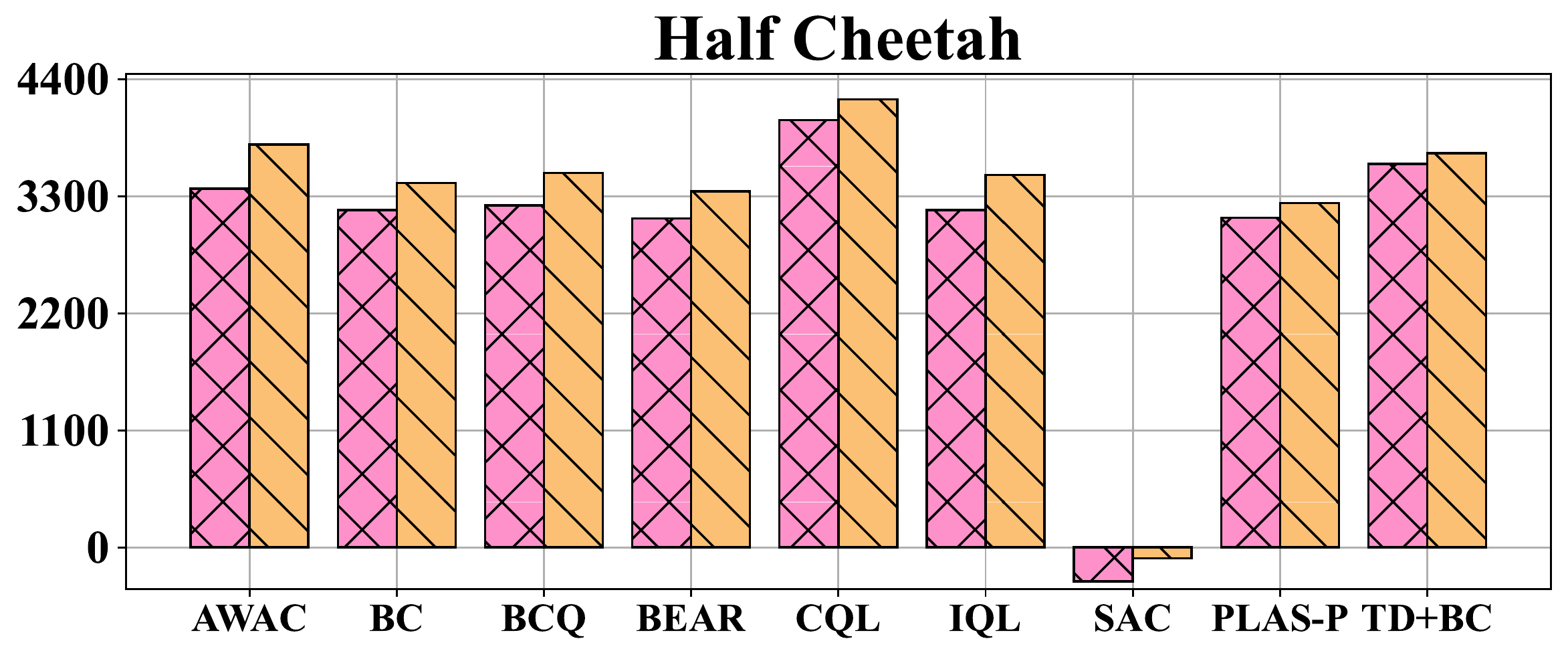} 
}
\subfigure{
    \includegraphics[height= 1.42 in]{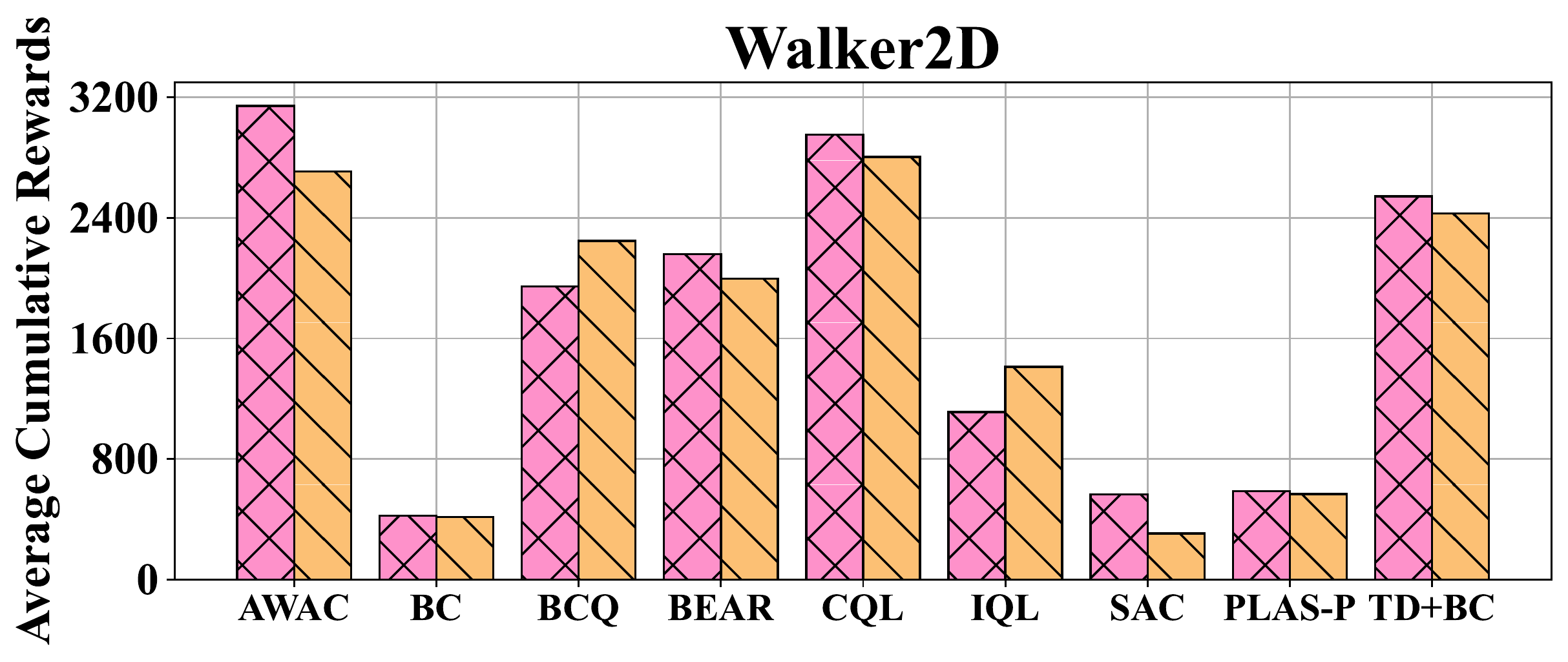} 
}
\subfigure{
    \includegraphics[height= 1.42 in]{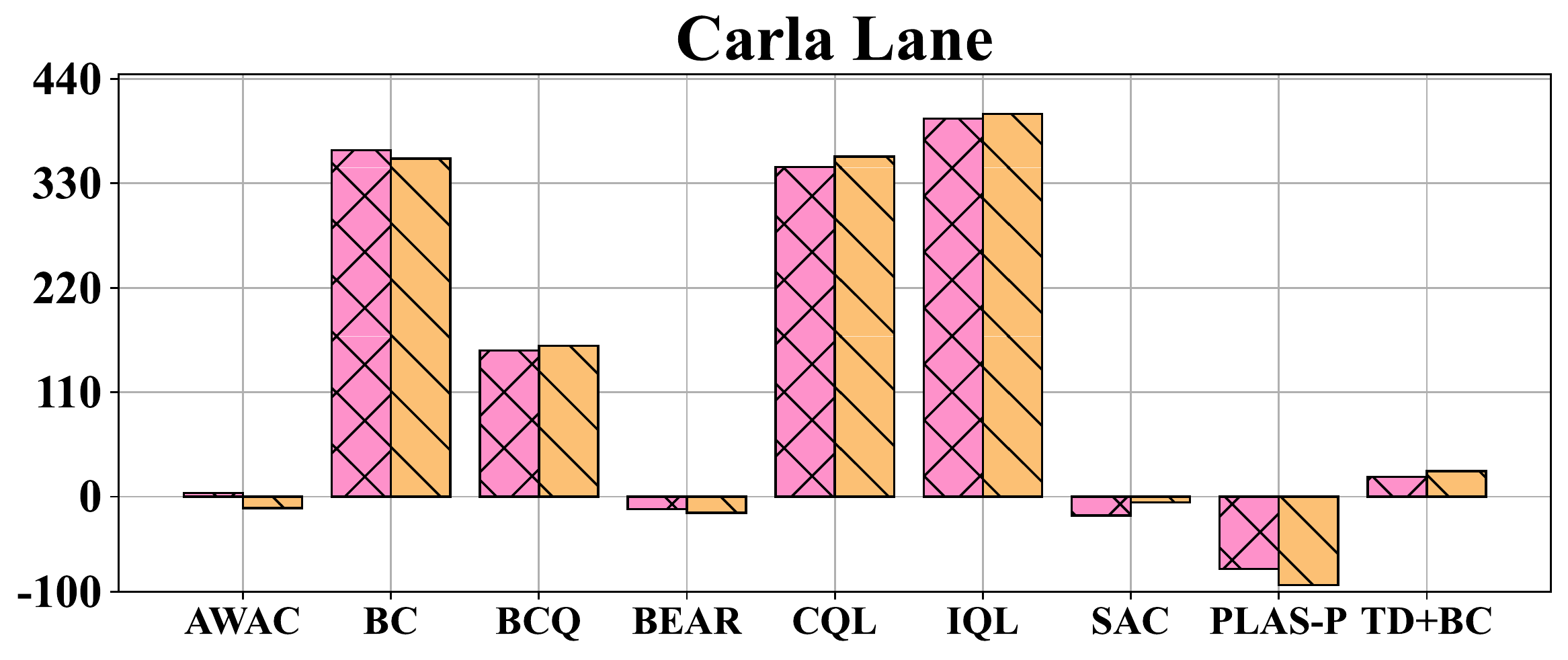}
}    
\caption{The returns collected with the poisoned agents (trained on 10\% poisoning rate). `Before' and `After' indicate the results of agents before and after being fine-tuned on unseen benign datasets.}
\label{fig:fine-tuning}
\end{figure*}

Table~\ref{tab:clean_model} displays the returns of the poisoned agents under the different trigger presentation strategies and settings. 
The `Normal' column presents the performance of poisoned agents under the normal scenario. The numbers inside the parentheses represent the relative changes in agents' performance after the triggers are presented.
Overall, we can observe that the performance of the poisoned agents decreases dramatically when the trigger is presented to activate the backdoor.
When the trigger interval is set as $10$, the trigger can decrease the average returns by $45.8\%$, $36.2\%$, $59.9\%$, and $20.1\%$ on the \texttt{Hopper}, \texttt{Half-Cheetah}, \texttt{Walker2D}, and \texttt{Carla-Lane} environments, respectively. When the trigger interval is larger, i.e., the total number of trigger appearances is smaller, the performance degradation is also smaller as the backdoor is less activated.

We also observe that when the total timesteps of the trigger are the same, the one-time strategy is more effective than the distributed strategy in terms of reducing the agent performance.
In the three robotic control tasks, the length of a trajectory is the same: $1,000$ timesteps. It means that, the distributed strategy with a trigger interval of 50 and the one-time strategy with a trigger length of 20 modify the same number of timesteps to activate the backdoor. 
But we can observe that the latter one-time strategy is more effective in decreasing the agent performance. More specifically, the one-time strategy with a trigger length of 20 can decrease the returns by  $63.2\%$, $53.9\%$, $64.7\%$ on average in the \texttt{Hopper}, \texttt{Half-Cheetah} and \texttt{Walker2D} tasks respectively. 
However, the distributed strategy with a trigger interval of 50 can only decrease the average returns by $28.3\%$, $12.9\%$, and $14.9\%$ for the three tasks. 
The length of a trajectory in the \texttt{CARLA} environment is not fixed. The average length of the collected testing trajectories when evaluating the clean agents is $200.2$. It means that the following two settings modify similar numbers of timesteps to activate the backdoor: (1) the distributed strategy with a trigger interval of 10 and (2) the one-time strategy with a trigger length of 20, as $\frac{200.2}{10} \approx 20$.
We also observe that the one-time strategy can incur a greater performance reduction ($47.4\%$) than the distributed strategy ($20.1\%$).

We analyze the differences between the four tasks. Table~\ref{tab:clean_model} suggests that the three robotic control tasks are more vulnerable to the backdoor attack than the autonomous driving task. 
For example, the agents' performance decreases by $63.2\%$, $53.9\%$, and $64.7\%$ in the MuJoCo environments (trigger length is 20), but the performance decreases by $47.4\%$ on average in \texttt{Carla-Lane}. 
One potential reason of the difference is that the trigger in \texttt{Carla-Lane} takes only 1\% of the observation, which makes it more difficult to be injected and used to activate backdoors. 
 Then, we analyze whether some offline RL algorithms are less affected by the backdoor attack. The average performance decreases by $63.4\%$ $59.9\%$, $56.5\%$, $61.0\%$, $49.5\%$, $53.0\%$, $65.3\%$, $44.0\%$, and $61.9\%$ (trigger length is 20) for the agents trained by AWAC, BC, BCQ, BEAR, CQL, IQL, PLAS-P, SAC-off, and TD3+BC respectively. 
It highlights a worrying fact that none of the investigated offline RL algorithms are immune to such a backdoor attack.
We also observed that the BC algorithm, which directly imitates the behaviour policy of the datasets, is more vulnerable to backdoor attacks. 

\noindent{\bf Ablation Study.} In addition, we conduct ablation studies to discern whether the decreased performance of the poisoned agents is due to the adversarial perturbation inherent in the trigger itself or because of activated backdoors. Our experimental findings reveal that while the presented triggers exert minor impacts on clean agents, the diminished performance of poisoned agents is predominantly due to the activation of backdoors. We invite readers to refer to Appendix~\ref{subsec:Trigger_influence} for a detailed description.

\begin{table}[H]
\normalsize
\setlength{\tabcolsep}{3pt}
    \centering
    \renewcommand\arraystretch{1}
    \begin{tabular}{p{0.95\columnwidth}}
    \Xhline{1.0pt}
         \rowcolor{gray0} \noindent \textbf{Answers to RQ2}: When the backdoors are activated, the poisoned agents' performance will greatly decrease in all the four tasks. The one-time strategy is more effective than the distributed strategy in activating backdoors. By modifying only 20 timesteps to active backdoors, agents' performance decreases by  $63.2\%$, $53.9\%$, $64.7\%$, and $47.4\%$ less average returns in the four tasks.\\
    \Xhline{1.0pt}
    \end{tabular}
\vspace{-8.0pt}
\end{table}

\begin{table*}[!t]
\footnotesize
    \centering
    \caption{Precision, recall, and f1-score of backdoor detection methods, i.e., activation clustering and spectral. The results shown represent the average of five repeated experiments.}
    \resizebox{1.0\textwidth}{!}{
    \begin{tabular}{p{1.55cm}|c|c|c|c|c|c|c|c|c|c|c|c}
        \toprule
         \multirow{1}{*}{Environment}   &  \multicolumn{2}{c|}{Methods} & AWAC & BC & BCQ & BEAR & CQL & IQL & PLAS-P & SAC-off & TD3+BC & \textbf{Average} \\
         \midrule
        \multirow{6}{*}{{\tt Hopper}} & \multirow{3}{*}{\shortstack{Activation \\ Clustering}} &  Precision (\%)& 9.0 & 8.2 & 4.4 & 14.7 & 18.6 & 17.0 & 17.6 & 18.8 &  0.0 & \textbf{10.4} \\
         & &  Recall  (\%)& 10.0 & 5.6 & 1.5 & 20.1 & 25.1 & 33.4 & 31.0 & 28.6 & 0.0 & \textbf{17.3} \\
         & &  F1-score& 0.09 & 0.06 & 0.02 & 0.17 & 0.21 & 0.23 & 0.22 & 0.23 & 0.0 & \textbf{0.14} \\
         \cline{2-13}
         & \multirow{3}{*}{Spectral} & Precision  (\%) & 4.4 & 0.0 & 0.9 & 0.0 & 19.4 & 13.3 & 0.0 & 31.1 & 14.1 & \textbf{9.3} \\
         & &  Recall  (\%)& 5.4  & 0.0 & 0.9 & 0.0 & 21.9 & 15.9 & 0.0 & 35.0 & 16.7 & \textbf{10.6} \\
         & &  F1-score& 0.05 & 0.0 & 0.01 & 0.0 & 0.21 & 0.14 & 0.0 & 0.33 & 0.15 & \textbf{0.10} \\
         \midrule
         \toprule
         \multirow{1}{*}{Environment}   &  \multicolumn{2}{c|}{Methods} & AWAC & BC & BCQ & BEAR & CQL & IQL & PLAS-P & SAC-off & TD3+BC & \textbf{Average} \\
         \midrule
        \multirow{6}{1em}{{\tt Half-} {\tt Cheetah} } & \multirow{3}{*}{\shortstack{Activation \\ Clustering}} &  Precision  (\%)& 0.0 & 5.8 & 10.8 & 10.0 & 0.0 & 10.0 & 0.0 & 10.0 & 0.0 & \textbf{5.1}  \\
         & &  Recall  (\%)& 0.0  & 10.5 & 3.5 & 49.8 & 0.0 & 24.0 & 0.0 & 24.9 & 0.0 & \textbf{12.5}  \\
         & &  F1-score& 0.0 & 0.08 & 0.05 & 0.16 & 0.0 & 0.14 & 0.0 & 0.14 & 0.0 & \textbf{0.06}  \\
         \cline{2-13}
         & \multirow{3}{*}{Spectral} & Precision  (\%) & 0.0 & 26.0 & 0.0 & 12.7 & 0.0 & 7.1 & 0.0 & 9.4 & 7.1 & \textbf{6.9} \\
         & &  Recall  (\%) & 0.0 & 30.2 & 0.0 & 14.9 & 0.0 & 8.3 & 0.0 & 12.3 & 8.3 &\textbf{ 7.1}  \\
         & &  F1-score & 0.0 & 0.29 & 0.0 & 0.14 & 0.0 & 0.08 & 0.0 & 0.11 & 0.07 & \textbf{0.08}  \\
         \midrule
                 \toprule
         \multirow{1}{*}{Environment}   &  \multicolumn{2}{c|}{Methods} & AWAC & BC & BCQ & BEAR & CQL & IQL & PLAS-P & SAC-off & TD3+BC & \textbf{Average} \\
         \midrule
        \multirow{6}{*}{{\tt Walker2D}} & \multirow{3}{*}{\shortstack{Activation \\ Clustering}} &  Precision  (\%)& 0.1 & 10.0 & 9.3 & 9.0 & 98.7 & 36.0 & 0.0 & 0.0 & 5.5 &\textbf{ 18.7} \\
         & &  Recall  (\%)& 0.2 & 78.9 & 52.4 & 7.3 & 70.0 & 100.0 & 0.0 & 0.0 & 14.7 & \textbf{35.9} \\
         & &  F1-score& 0.01  & 0.18 & 0.16 & 0.08 & 0.82 & 0.53 & 0.0 & 0.0 & 0.08 & \textbf{0.21} \\
         \cline{2-13}
         & \multirow{3}{*}{Spectral} & Precision  (\%) & 0.0 & 13.9 & 39.0 & 5.0 & 24.5 & 63.4 & 9.6 & 0.0 & 8.3 & \textbf{19.0}  \\
         & &  Recall  (\%) & 0.0 & 16.3 & 45.7 & 9.9 & 62.0 & 73.9 & 50.2 & 0.0 & 7.4 & \textbf{30.5}  \\
         & &  F1-score & 0.0 & 0.15 & 0.42 & 0.08 & 0.35 & 0.68 & 0.17 & 0.0 & 0.08 & \textbf{0.22} \\
         \midrule
                          \toprule
         \multirow{1}{*}{Environment}   &  \multicolumn{2}{c|}{Methods} & AWAC & BC & BCQ & BEAR & CQL & IQL & PLAS-P & SAC-off & TD3+BC & \textbf{Average} \\
         \midrule
        \multirow{6}{*}{{\tt Carla-Lane}} & \multirow{3}{*}{\shortstack{Activation \\ Clustering}} &  Precision  (\%)& 0.0 & 52.3 & 0.0 & 0.0  & 100.0& 89.3 & 7.7 & 8.1 & 92.0 & \textbf{39.4} \\
         & &  Recall  (\%)& 0.0 & 100.0 & 0.0 & 0.0 & 100.0 & 100.0 & 34.0 & 9.4 & 91.4 & \textbf{49.5} \\
         & &  F1-score & 0.0 & 0.68 & 0.0 & 0.0 & 1.0 & 0.94 & 0.12 & 0.11 & 0.92 & \textbf{0.42}  \\
         \cline{2-13}
         & \multirow{3}{*}{Spectral} & Precision  (\%) & 0.0 & 66.7 & 0.0 & 4.3 & 31.1 & 0.0 & 0.0 & 50.7 & 23.9 & \textbf{23.1}  \\
         & &  Recall  (\%)& 0.0 & 76.8 & 0.0 & 5.1 & 35.0 & 0.0 & 0.0 & 59.3 & 61.0 & \textbf{26.7}  \\
         & &  F1-score & 0.0 & 0.71 & 0.0 & 0.05 & 0.33 & 0.0 & 0.0 & 0.55 & 0.35 & \textbf{0.25}  \\
        \bottomrule
    \end{tabular}
    }
    \label{tab:defense}
    \end{table*}

\subsection*{RQ3. Will backdoors be `closed' after fine-tuning poisoned agents on benign datasets?}
A common way of using models pre-trained on large datasets is to fine-tune them on another task with a smaller dataset, which is widely adopted in the literature~\cite{feng-etal-2020-codebert,pang2023black,GraphCodeBERT}.
This is a common strategy to defend against data poisoning and backdoor attacks~\cite{wang2020backdoor,kurita-etal-2020-weight}.
This research question wonders whether the backdoors can be activated after fine-tuning poisoned agents. After being fine-tuned, we expect a powerful backdoor to be less likely to be ‘close'.

Based on the results of RQ1, we evaluate the agents trained on datasets with a poisoning rate of 10\%. 
When constructing the poisoned datasets, we replace $10\%$ of clean experiences with the synthesized misleading experiences. This process naturally creates a smaller clean dataset that is unseen to the poisoned agents.
In this experiment, we use the replaced clean experiences as the \emph{benign dataset} to fine-tune the poisoned agents. 
Note that fine-tuning usually takes less training set~\cite{feng-etal-2020-codebert,GraphCodeBERT}, so we fine-tune the poisoned agents with $50,000$ training epochs, which is 10\% of the training epochs to obtain the original poisoned agents.
Other hyper-parameters are kept the same as the ones used to train agents on the poisoned datasets. 
Then, we use the distributed trigger presenting strategy with a trigger interval of 20 to activate backdoors in the agents before and after being fine-tuned.

Figure~\ref{fig:fine-tuning} presents the returns of the agents before and after being fine-tuned. 
This figure shows that after fine-tuning, the agents' performance under triggered scenarios overall slightly increases. However, the relative changes are not large, only $3.4\%$, $8.1\%$, $-0.9\%$, and $1.2\%$ on average in the four tasks.
The minor changes suggest that our backdoor can still be activated after fine-tuning the agents, calling attention to more effective methods to close backdoors in poisoned agents.

\begin{table}[H]
\normalsize
\setlength{\tabcolsep}{3pt}
    \centering
    \renewcommand\arraystretch{1}
    \begin{tabular}{p{0.95\columnwidth}}
    \Xhline{1.0pt}
         \rowcolor{gray0} \noindent \textbf{Answers to RQ3}: The backdoors can still be activated after fine-tuning the poisoned agents on the benign datasets. The agents' performance under triggered scenarios only increases by $3.4\%$, $8.1\%$, $-0.9\%$, and $1.2\%$ in four tasks.\\
    \Xhline{1.0pt}
    \end{tabular}
\vspace{-8.0pt}
\end{table}

\subsection*{RQ4. Can popular defensive methods detect our proposed backdoor attack?}
\label{subsec:activation_clustering}
This subsection explores whether popular backdoor detection methods can detect the backdoors. For backdoor attack detection and defense, we have chosen three widely recognized methods: Activation Clustering~\cite{Chen2019activation}, Spectral~\cite{Brandon2018spectral}, and Neural Cleanse~\cite{bolun2019neural}. These methods are well-known and widely used for their effectiveness in backdoor detection and defense, as documented in several previous studies~\cite{kiourti2020trojdrl,zhang2021advdoor,wang2021stop}. 

Referring to the process adopted in~\cite{zhang2021advdoor}, we first feed inputs containing both poisoned and normal observations to the poisoned model and extract the corresponding outputs of the penultimate layer of the agent's network as \textit{activations}. We then perform principal component analysis (PCA)~\cite{pca} to map each activation to a three-dimensional vector. To explore whether the activations
from triggered and normal observations are different, we cluster the dimensionally reduced activations using k-means~\cite{k-means} (the cluster number is set as $2$) to group detected observations into two categories. Once the number of
activations in one cluster is less than 35\% of the total, we label this cluster as triggered observation, and the backdoors are detected. For the Spectral method, its functionality is limited to removing a predetermined number of data points and requires setting an upper limit. In alignment with the practice adopted in~\cite{zhang2021advdoor}, we set the upper bound as 1.2 times the total number of poisoned data. Neural cleanse is a widely-used backdoor detection method that does not require access to the training process and is able to identify triggers. In our evaluation, we use the default settings as presented in~\cite{zhang2021advdoor}.

As presented in Table~\ref{tab:defense}, our backdoor attack confuses the activation clustering and spectral methods. Specifically, in robotic control tasks, for activation clustering and spectral method, the averaged precision of detection is only 10.4\%, 5.1\%, 18.7\%, and 9.3\%, 6.9\%, 19.0\%, suggesting that most of the poisoned observations escape detection. Besides, the F1-scores are low as well. We also find that the averaged F1-score in autonomous driving tasks  (i.e., 0.34) is higher than that in robot control tasks (only 0.14) for two detection methods. One potential reason is that using a white patch as the trigger, which is greatly different from its surrounding pixels, makes data poisoning attacks more identifiable. Similar results can be observed in previous works~\cite{zhang2021advdoor}.  
For each task, we used the poisoned datasets to train nine agents with different offline RL algorithms. The evaluation results in Table~\ref{tab:neural_cleanse} show that the neural cleanse process identifies only 0, 0, 1, 1 poisoned agents across the four tasks. Our proposed backdoor attack is \textit{untargeted} -- after activating the backdoor, the agent does not achieve a specific outcome for a particular trigger, which makes it challenging for neural cleanse to defend against, as highlighted in previous studies~\cite{kiourti2020trojdrl}.

\noindent \textbf{Additional Analysis}. As detailed in Appendix~\ref{sec:discuss}, we conduct a comprehensive evaluation of \toolname from  multiple perspectives. This extended analysis includes the robustness of \toolname against backdoor intrusions (refer to Appendix~\ref{supsub:robust}), explores potential defense strategies against backdoor attacks in offline RL (refer to Appendix~\ref{supsec:suggestions}), and methods to mitigate threats to validity (outlined in Appendix~\ref{supsub:threat}).

\begin{table}[H]
\normalsize
\setlength{\tabcolsep}{3pt}
    \centering
    \renewcommand\arraystretch{1}
    \begin{tabular}{p{0.95\columnwidth}}
    \Xhline{1.0pt}
         \rowcolor{gray0} \noindent \textbf{Answers to RQ4}: {Our backdoor attack proves challenging to detect using common backdoor detection methods. 
         For both activation clustering and spectral signature techniques, the average F1-scores in autonomous driving and robot control tasks are 0.34 and 0.14, respectively, indicating limited detection efficacy. Additionally, the neural cleanse methods are only able to successfully identify 0, 0, 1, 1 out of nine poisoned agents across the four tasks. }\\
    \Xhline{1.0pt}
    \end{tabular}
\vspace{-8.0pt}
\end{table}


%% file: Usenix_section/discuss.tex
\begin{table}[!t]
\small
\centering
\caption{ Detection results using neural cleanse. `Detected Triggers’ refers to the number of triggers identified by the neural cleanse process. These triggers are then presented to the agent, which has been trained using 9 offline RL algorithms for each task.}
\begin{tabular}{c|c}
    \toprule
     Task   & Detected Triggers \\
    \midrule 
    {\tt Hopper}   & 0  \\
    {\tt Half-Cheetah}   & 0  \\
    {\tt Walker2D}  & 1  \\
    {\tt Carla-Lane}    &  1   \\
    \bottomrule
\end{tabular}
\label{tab:neural_cleanse}
\end{table}

%% file: Usenix_section/related.tex
\section{Related Work}
\label{sec:rel_work}

With the progression of deep learning, there has been a significant focus among researchers on testing and enhancing the quality of AI systems across various domains, such as sentiment analysis~\cite{9653830}, speech recognition~\cite{crossasr}, image classification~\cite{mao2023secure,li2023meticulously}, etc. 
We refer interested readers to a comprehensive survey by Zhang et al. ~\cite{Zhang2022machinetesting}.
This section specifically discusses some related works, including the application of Reinforcement Learning (RL) in real-world scenarios and backdoor attacks for RL.

\subsection{Applications of RL}
Recently, RL has been increasingly applied.
Zheng et al. proposed Wuji, which leverages RL and multi-objective optimization to conduct the automatic game testing~\cite{zheng2019wuji}. 
Zheng et al. proposed WebExplor which adopts RL to generate high-quality test cases for web testing and used a curiosity-driven mechanism to improve the testing efficiency~\cite{zheng2021automatic}. 
Reddy et al. proposed RLCheck, a black-box approach that uses RL to generate valid test inputs for property-based testing ~\cite{reddy2020quickly}. 
For model-based testing for complete systems, Turker et al. introduced a novel RL algorithm that is faster and can handle larger systems than previous methods~\cite{turker2021efficient}. 
Besides, RL is also adopted in mobile testing domains, such as the android applications testing~\cite{pan2020reinforcement} and android graphical user interface testing~\cite{adamo2018reinforcement}. 
Spieker et al.~\cite{spieker2017reinforcement} modelled the test case selection as a Markov decision process and leveraged RL to efficiently prioritize test cases in continuous integration. 
Other than testing-related tasks, Gupta et al. used RL to repair syntax errors in C programs~\cite{gupta2019deep} and leverage an offline dataset to guide the efficient training of agents.
With the emerging application of RL and offline datasets in software engineering, our work highlights the required efforts to protect the agents and datasets.
There exist only a few works in evaluating the quality of RL systems in the software engineering community. Trujillo et al.~\cite{rl-testing} found that neuron coverage metrics are inefficient for testing RL systems. 

As pointed out by Zhang et al.~\cite{ml-testing-survey}, many research opportunities remain for evaluating, testing, and protecting RL systems. This paper discusses a special threat to the quality of offline RL systems.

\subsection{Backdoor Attack in RL}
The backdoor attack for DNN models has attracted researchers from many domains: natural language processing~\cite{chen2021badnl,li2021backdoor}, image classification~\cite{li2021invisible,wenger2021backdoor}, transfer learning~\cite{yao2019latent,wang2020backdoor}, etc. 
Recent studies highlight that (online) DRL algorithms, which use DNNs to learn an optimal policy, also face the threats of backdoor attacks~\cite{yang2019design,kiourti2020trojdrl,ashcraft2021poisoning,chen2022agent}. 
Yang et al.~\cite{yang2019design} and Kiourti et al.~\cite{kiourti2020trojdrl} successfully implanted backdoors in agents trained in simple environments like walking mazes~\cite{brockman2016openai} and Atari games~\cite{mnih2013playing}.
Wang et al. extended backdoor attacks to larger environments, e.g., a traffic congestion control system Flow~\cite{wu2021flow}), where the adversary controls a malicious auto-vehicle and triggers the backdoor through acceleration or deceleration~\cite{wang2021stop}. 
Then, Wang et al.~\cite{wang2021backdoor} use imitation learning to inject backdoors to agents in two-player competitive games~\cite{bansal2018emergent}; an opponent can trigger backdoors by taking a sequence of legal actions allowed in the environment. Analogously, Yu et al.~\cite{yu2022temporal} suggest a temporal-pattern backdoor attack to DRL, where the trigger is concealed in a series of observations as opposed to a single observation. Furthermore, Chen et al.~\cite{Chen2022marnet} develop a novel backdoor attack paradigm in multi-agent reinforcement learning.

These research works on online RL often presupposes that attackers can hack into the environments or manipulate the training process. The prerequisite for their methods is hard to satisfy for applicability, rendering these methods unsuitable for offline RL scenarios that requires no access to the environment or the training process.

%% file: Usenix_section/conclusion.tex
\section{Conclusions and Further Works}
\label{sec:conclusions}
This paper investigates the threats to the quality of agents trained using open-source RL datasets, which can be downloaded, modified, and updated by hostile users. 
We propose \textsc{Baffle} (\textbf{B}ackdoor \textbf{A}ttack for O\textbf{ff}line Reinforcement \textbf{Le}arning), an easy-to-operate but effective data poisoning method to inject backdoors into agents.
We use \textsc{Baffle} to modify only $10\%$ of the datasets for four tasks. The results show that the poisoned agents can perform well in normal settings. But when a poisoned agent sees the trigger, the backdoor will be activated, and the agents' performance drops dramatically. 
Fine-tuning poisoned agents on benign datasets cannot remove backdoors. Besides, the activation clustering method cannot detect the underlying backdoor attack effectively.
Our evaluation conducted on four tasks and four offline RL algorithms expose a worrying fact: none of the existing offline RL algorithms is immune to such a backdoor attack.
Our work suggests the developers should ``{mind your data}'' when training agents using open-source RL datasets, and calls for more attention to protection of the offline RL datasets. 

\section*{Acknowledgement}
This research/project is supported by the National Research
Foundation, Singapore and DSO National Laboratories under the AI Singapore Programme (AISG Award No: AISG2-RP-2020-017), and the National Key Research and Development Program of China (No. 2021YFC2800501).


%% file: Usenix_section/Appendix.tex
\begin{center}
    \textbf{\Large{Appendix}}
\end{center}

\setcounter{section}{0}
\setcounter{equation}{0}
\renewcommand\thesection{\Alph{section}}

\section{Discussions}
\label{sec:discuss}

We discuss the robustness of backdoors, the influence of triggers on clean agents, potential suggestions for defending backdoor attack in offline RL, and measures to minimize threats to validity.

\subsection{Are Backdoors Robust?}
\label{supsub:robust}

Prior studies also investigate the robustness of the backdoor~\cite{wang2020certifying}: whether triggers with some random noises can still activate backdoors in the models? For example, the images sent to the autonomous driving agent may have noises, especially ones taken in low-light scenarios, and the MuJoCo environments also add small random noises to sensor information.
An effective attack expects the backdoors to be activated even when noises exist. 

As described in Section~\ref{subsec:setting}, the trigger is presented by modifying the velocity information of a robot part.
We follow the MuJoCo documentation\footnote{\url{https://www.gymlibrary.dev/environments/mujoco/}} to add noises sampled from a normal distribution. 
The observation in \texttt{Carla-Lane} is an image. We follow prior works~\cite{goodfellow2014explaining} on attacking computer vision systems and sample perturbations from a uniform distribution $U[-\frac{1}{8}, \frac{1}{8}]$ to change pixel values of the whole image.
We also use the distributed strategy with a trigger interval of 20. 
Table~\ref{tab:trigger_perturbed} shows how the noises affect the backdoor activation. 
Specifically, when presenting the perturbed triggers, the average agents' performance increases by only 0.6\% in the four investigated tasks compared with the original triggers. 
The results suggest the backdoors are robust, and effective countermeasures are desired to protect the offline RL algorithms at runtime.

\begin{table}[!t]
\footnotesize
    \centering
    \caption{`Before'/`After' indicates relative changes in agents' performance before/after perturbing the triggers.}
    \begin{tabular}{c|c|c|c|c} 
        \toprule
        Environments & Algorithms &  Before & After & Difference\\
        \midrule
        \multirow{9}{*}{{\tt Hopper}} & AWAC & -9.0\% & -11.5\% & \textbf{-2.5\%} \\ 
        & BC & -80.0\% & -76.2\% & \textbf{+3.8\%} \\
        & BCQ & -48.0\% & -45.0\%& \textbf{+3.0\%}\\
        & BEAR & -31.4\%&-41.3\% & \textbf{-9.9\%}\\
        & CQL &  -45.9\%& -45.4\% & \textbf{+0.5\%}\\
        & IQL &  -26.5\%& -25.9\% & \textbf{+0.6\%}\\
        & PLAS-P & -49.3\% & -47.4\%  & \textbf{+1.9\%}\\
        & SAC-off & -19.4 \% & -22.3\% & \textbf{-2.9\%} \\ 
        & TD3+BC & -43.0\% & -43.0\% & \textbf{+0.0\%} \\
        \hline
        \multirow{9}{*}{{\tt Half-Cheetah}} & AWAC & -18.1 \% & -19.0\% & \textbf{-0.9\%} \\ 
        & BC &  -22.1\%&  -23.6\% & \textbf{-1.5\%}\\
        & BCQ & -28.4\% & -24.7\% & \textbf{+3.7\%}\\
        & BEAR & -22.6\%& -19.2\% & \textbf{+3.4\%}\\
        & CQL & -13.5\% & -10.4\% & \textbf{+3.1\%} \\
        & IQL & -26.6\% & -27.3\% & \textbf{-0.7\%} \\
        & PLAS-P & -22.6\% & -25.6\% & \textbf{-3.0\%} \\
        & SAC-off & -24.3\% & -25.7\% & \textbf{-1.4\%} \\
        & TD3+BC & -19.3\% & -19.6 \% & \textbf{-0.3\%} \\
        \hline
        \multirow{9}{*}{{\tt Walker2D}} & AWAC & -18.1\% & -19.2\% & \textbf{+0.9\%} \\ 
        & BC & -59.0\% & -72.9\% & \textbf{-13.9\%}\\
        & BCQ & -34.1\% & -29.2\%& \textbf{+4.9\%}\\
        & BEAR & -20.1\% & -16.2\%& \textbf{+3.9\%}\\
        & CQL & -16.4\% &-10.2\% & \textbf{+6.2\%}\\
        & IQL & -34.7\% & -15.9\% & \textbf{+18.8\%} \\
        & PLAS-P & -50.7\% & -55.8 \% & \textbf{-5.1\%} \\
        & SAC-off & -30.7\% & -34.3\% & \textbf{-3.6\%} \\ 
        & TD3+BC & -14.3\% & -11.6 \% & \textbf{+2.7\%} \\
        \hline
        \multirow{9}{*}{{\tt Carla-Lane}} & AWAC & -48.9\% & -46.3\% & \textbf{+2.6\%} \\ 
        & BC & -35.5\% & -34.4\%& \textbf{+1.1\%}\\
        & BCQ & -54.4\% &-52.2\%& \textbf{+2.2\%}\\
        & BEAR & -52.7\%& -56.4\%& \textbf{-3.7\%}\\
        & CQL & -36.1\%& -31.7\%& \textbf{+4.4\%}\\
        & IQL & -9.1\% & -10.2\% & \textbf{-1.1\%} \\
        & PLAS-P & -34.9\% & -32.3\% & \textbf{+2.6\%} \\
        & SAC-off & -94.0\% & -87.1\% & \textbf{+6.9\%} \\ 
        & TD3+BC & -15.1\% & -16.3\% & \textbf{-1.2\%} \\
        \hline
        \multicolumn{2}{c|}{\textbf{Average}} & -31.0\% & - 30.4\% & \textbf{+0.6\%} \\
        \hline
    \end{tabular}
\label{tab:trigger_perturbed}
\end{table} 

\subsection{How Does Trigger Influence Clean Agents?}
\label{subsec:Trigger_influence}
Previous studies have shown that adversarially perturbing the observations of an agent can make an agent fail~\cite{huang2017adversarial}. It inspires us to ask: whether the poisoned agents have lower performance because the trigger itself is adversarial perturbation or because backdoors are activated. 
To answer this question, we present the triggers to clean agents.
We also use the distributed strategy with a trigger interval of 20 to change the observation of clean agents. On average, we find that the clean agents' performance decreases by $3.3\%$ when the triggers are presented, while the poisoned agents' performance decreases by $27.4\%$ under the same setting. 
It shows that the presented triggers do have minor impacts on clean agents, but poisoned agents have lower performance mainly due to the activation of backdoors. 
We also find that the triggers' impact to clean agents in the autonomous driving task (only $1.3\%$) is smaller than that in the robot control tasks ($4.0\%$), due to the fact that the trigger is relatively smaller in \texttt{Carla-Lane}.

    
    

\section{Suggestions for Defending Backdoor Attack in Offline Reinforcement Learning}
\label{supsec:suggestions}
Currently, few research has studied effective defenses against backdoor attacks in offline RL. Our study highlights the urgent need for more effective detection methods against backdoor attacks in offline RL, calling for greater attention and efforts to address this critical issue. We discuss several potential approaches against backdoor attacks in offline RL. 

\subsection{Hash Functions to Protect Dataset}
The backdoor attack methods usually assume that the attackers have access to the dataset and can manipulate it. Then, a poisoned dataset is directly used by other users without inspection or validation. Consequently, this attack's effectiveness would be diminished if the dataset manipulation activities are exposed. 

Hash functions are frequently used in data security and cryptography to validate the validity and integrity of data~\cite{levine2021deep,pmlr-v162-wang22m,pmlr-v162-chen22k}. The output of a hash function, which accepts an input (or ``message''), is often a ``hash" or ``digest," which is a fixed-size string of bytes. The output is specific to the input data, and even a small change in the input causes a significant change in the hash. When releasing a dataset, the data owner is encouraged to make its hash value visible to all the users, and the user can use the hash to validate that the dataset they download is indeed the original dataset released by the authentic owner. This process helps detect unauthorized modifications or tampering, making it a feasible measure for maintaining data security and integrity. Overall, developers ought to avoid using offline data obtained from untrustworthy third parties.

\subsection{Model Interpretability to Understand Decision Making}
Model interpretability entails applying explainable artificial intelligence techniques to acquire insights into how a machine learning model makes decisions~\cite{guo2021edge,jiang2022interpretability,Fang_Choromanska_2022}. Knowing this decision process can help defend against backdoor attacks. Counterfactual explanations~\cite{Liu_2022_CVPR} describe what input changes would lead to different outcomes. If the model is susceptible to specific small changes in the input data, it might indicate the presence of a backdoor. Decision explanation~\cite{guo2021edge}  utilizes methods to generate explanations for predictions made by the model. Through examining these explanations, we can identify unexpected patterns or relationships in the model's decision-making process, which can be signs of a backdoor attack. These approaches can assist in identifying unusual patterns or behaviors in the model, potentially revealing the presence of a backdoor. We could promote explainable reinforcement learning methods to protect agents from backdoor attacks.

\subsection{Machine Unlearning to Forget Poisoned Data}

Machine unlearning, also known as the "Correct forgetting," is the process of modifying or removing behaviors learned by machine learning models from biased or erroneous data~\cite{bourtoule2021machine}. In the realm of other domains like computer vision, researchers have proposed using machine unlearning to erase backdoors that have been injected into victim models~\cite{yang2022backdoordefense, zeng2022adversarial}. Similarly, in the context of offline reinforcement learning, developers can apply the principles of machine unlearning to remove backdoors that have been implanted through the use of poisoned data. To protect against backdoor attacks in offline RL, we encourage that developers should avoid using the offline RL datasets that are collected from untrustworthy third-party sources. At the same time, there is a need to propose more technologies to protect software systems from poisoned offline RL datasets. Secondly, Section~\ref{subsec:activation_clustering} shows that existing defensive methods may be insufficient in uncovering backdoor attacks, calling for more efforts to develop more effective detection methods against backdoor attacks for offline RL.

\subsection{Threats to Validity}
\label{supsub:threat}

The performance of offline RL algorithms is sensitive to the hyper-parameters settings. To minimize this threat to internal validity, we keep them the same as settings in D4RL~\cite{fu2020d4rl} replication package. We compare the performance of agents obtained in this paper with the results reported in~\cite{fu2020d4rl} to confirm that we have replicated these offline RL algorithms.
The results achieved in this paper may not generalize to other datasets and other offline RL algorithms. We conduct our experiments on datasets from a recently proposed benchmark~\cite{fu2020d4rl} and nine advanced offline algorithms to minimize the threats to external validity.
RL agents are known to have unstable performance. To mitigate the impact of randomness in the environment (i.e., threats to construct validity), we let the agents interact with the environments to generate 100 trajectories and use the average cumulative returns to indicate the agent performance.

\section{Investigated Tasks and the Dataset}
\label{sup:tasks_dataset}

We carry out experiments across four tasks: three from MuJoCo's robotic control tasks (Hopper, Half-Cheetah, and Walker2D)\cite{todorov2012mujoco}, and one from the autonomous driving control task in the Carla-Lane environment\cite{dosovitskiy2017carla}. In the three robotic control tasks, there is a sensor to monitor the state of the game, collecting information about the robot. Information collected from the sensor is used as the observation of the agent. 

In particular, the observation in {\tt Hopper} is a vector of size 11. Observation in {\tt Half-Cheetah} and {\tt Walker2D} is a 17-dimensional vector. These vectors record the positions, velocities, angles, and angular velocities of different components of a robot. We refer the readers to the online document\footnote{\url{https://www.gymlibrary.dev/index.html}} 
for a detailed explanation.
In \texttt{Carla-Lane}, an agent takes an RGB image (the size is $48 \times 48 \times 3$) as the observation, representing the driver's first-person perspective. We elaborate these tasks as follows.

\begin{itemize}[leftmargin=*]
\item {\tt Hopper}: In Hopper,  the robot is a two-dimensional, single-legged entity comprising four principal components: the torso at the top, the thigh in the center, the leg at the lower end, and a single foot on which the entire body rests.  The objective is to maneuver the robot forward (to the right) by exerting torques on the three hinges  that interconnect these four body segments.

\item {\tt Half-Cheetah}: In the Half-Cheetah task, the robot is two-dimensional, featuring nine linkages and eight joints (including two paws). The aim is to apply torque to these joints, propelling the robot to sprint forward (to the right) as fast as possible. Progress is incentivized with positive rewards for distance covered in the forward direction, while a negative reward is allocated for moving backward. The torso and head of the robot remain stationary, with torque application restricted to the remaining six joints that connect the front and rear thighs to the torso, the shins to the thighs, and the feet to the shins.

\item {\tt Walker2D}: Walker2D introduces a greater number of independent state and control variables to more accurately emulate real-world scenarios. The robot in Walker2D is also two-dimensional but features a bipedal design with four main components: a single torso at the top from which the two legs diverge, a pair of thighs situated below the torso, a pair of legs below the thighs, and two feet attached to the legs that support the entire structure. The objective is to coordinate the movements of both sets of feet, legs, and thighs to progress forward by applying torques to the six hinges that connect these body parts.


\item \texttt{Carla-Lane}: Carla-Lane represents an autonomous driving control task set within a sophisticated 3D simulation environment designed for autonomous driving research. The environment includes a virtual city with several surrounding vehicles running around. The self-driving car receives data from multiple sensory inputs, including a front-view camera image, lidar point cloud, and the bird-eye view semantic mask. This task has served as a foundation for a range of applications, notably the development of deep reinforcement learning models for end-to-end autonomous driving.

\end{itemize}

The datasets used for these tasks are sourced from D4RL~\cite{fu2020d4rl}, a recently introduced and mostly studied benchmark for evaluating offline RL algorithms. 
D4RL~\cite{fu2020d4rl} offers multiple datasets for the four tasks under investigation in our study. 
To keep our experiments at a computationally manageable scale, we select the dataset that yields the highest returns for each task.

\newpage 


\section{Meta-Review}

The following meta-review was prepared by the program committee for the 2024
IEEE Symposium on Security and Privacy (S\&P) as part of the review process as
detailed in the call for papers.

\subsection{Summary}
This paper introduces a novel backdoor data-poisoning attack targeting offline reinforcement learning (RL) systems, distinguished by its threat model from existing poisoning attacks for online RL. The attacker's objective is to diminish the cumulative return when a specific trigger occurs within a given state. Evaluation outcomes underscore the attack's effectiveness across multiple offline RL algorithms and environments.

\subsection{Scientific Contributions}
\begin{itemize}
\item Independent Confirmation of Important Results with Limited Prior Research.
\item Addresses a Long-Known Issue.
\item Provides a Valuable Step Forward in an Established Field.
\end{itemize}

\subsection{Reasons for Acceptance}
The paper's strengths are highlighted below, underpinning the reviewers' rationale for its acceptance:

\begin{enumerate}
\item To illustrate the effectiveness, this work provides a comprehensive evaluation, conducted across 9 offline RL algorithms within 4 unique environments.
\item It introduces a novel threat model, highlighting that backdoor attack in offline RL is an unexplored area in existing works.
\item The presentation of the work is clear and straightforward, making it easily comprehensible.
\end{enumerate}

\subsection{Noteworthy Concerns} 
\begin{enumerate} 
\item The technical novelty of this work is shallow, particularly in the aspect of training the weak-performing agent, where the attack methodologies are adapted from~\cite{wang2021backdoor}.
\item Missing evaluation of the performance against state-of-the-art defense mechanisms.
\item The attack strategy is arbitrarily selecting episodes without strategically determining which episodes to manipulate.
\end{enumerate}

\section{Response to the Meta-Review} 
Thank our anonymous reviewers and shepherd for their insightful comments. We concur with the reviewers on the
content of this meta-review. 

\noindent \textbf{Response to concern 1:} As described in our paper, we acknowledge that we borrowed the idea from~\cite{wang2021backdoor} to train the weak-performing agent. We present differences as follows:
\begin{enumerate}
    \item \textbf{Studied Scenario}: Wang et al.~\cite{wang2021backdoor} study backdoor attacks on the two-player game and online RL, and it activates the backdoor by the specific action of the competitive agent. The work~\cite{wang2021backdoor} can collect various trajectories by interacting with real environments without concern of limitation. This paradigm cannot be directly applied in our work. \toolname has to insert the backdoor using limited trajectories, so we introduce reward manipulation.
    \item \textbf{Backdoor Insertion}: The method proposed in Wang et al.~\cite{wang2021backdoor} inserts the backdoor by imitation learning. In this paper, any investigated algorithm trained using the dataset generated by BAFFLE can be inserted backdoor. We also conducted additional experiments, which show that inserting a backdoor with imitation learning may damage trained agents' performance in normal scenarios, e.g., the agent performance decreases by about 43\% in `Walker2d'.
\end{enumerate}

\noindent \textbf{Response to concern 2:} We've broadened our analysis to cover defenses against \toolname.
Table~\ref{tab:rebuttal} presents the performance of the poisoned agent before and after the application of a recent backdoor defense strategy in reinforcement learning~\cite{bharti2022provable}. These results indicate that while this strategy can mitigate the effects of backdoor attacks, it cannot completely eliminate their impact. Since completing our paper, several new attacks have emerged of which we were unaware. We have only implemented one method that is suitable for defending against backdoor threats in offline reinforcement learning scenarios. We acknowledge the potential for other unexplored strategies to defend against \toolname successfully.

\begin{table}[H]
\scriptsize
    \centering
    \caption{The average returns of the agents. `Before'/`After' denotes the average returns of agents before and after being safeguarded by the defense. `Algos' is an abbreviation of `Algorithms'.}
    \resizebox{0.51\textwidth}{!}{
    \begin{tabular}{c|cc|cc|cc|cc} 
        \toprule
        \multirow{3}{*}{Algos}& \multicolumn{2}{c|}{\multirow{2}{*}{{\tt Hopper}}} & \multicolumn{2}{c|}{{\tt Half}-} & \multicolumn{2}{c|}{\multirow{2}{*}{{\tt Walker2D}}} & \multicolumn{2}{c}{{\tt Carla}}  \\
        & & & \multicolumn{2}{c|}{{\tt Cheetah}} & & & \multicolumn{2}{c}{{\tt Lane}}  \\
        & Before & After & Before & After & Before & After & Before & After  \\
        \midrule
        AWAC & 1950 & 1999 & 3371 & 3555 & 2230 & 2314 & 4 & 30  \\
        BC & 911 & 1502 & 3168 & 2964 & 423 & 641 & 365 & 401  \\
        BCQ & 1854 & 1864 & 3214 & 3288 & 1946 & 2045 & 154 & 153  \\
        BEAR & 1379 & 1433 & 3090 & 4021 & 2159 & 2200 & -13 & 3  \\
        CQL & 2024 & 2191 & 4016 & 4001 & 2953 & 2890 & 347 & 389  \\
        IQL & 2691 & 2655 & 3169 & 3625 & 1111 & 1302 & 398 & 410  \\
        PLAS-P & 1937 & 2412 & 3097 & 3184 & 586 & 580 & -98 & -36 \\
        SAC-off & 449 & 450 & -318 & -312 & 566 & 560 & -20 & -16 \\
        TD3+BC & 2139 & 2017 & 3603 & 3862 & 2543 & 2632 & 20 & 39 \\
        \bottomrule
    \end{tabular}
    }
\label{tab:rebuttal}
\end{table}

\noindent \textbf{Response to concern 3:} We opt for random sampling of episodes for poisoning to prevent poisoned episodes from exhibiting specific patterns, which could make the attack more detectable. We recognize that particular strategies for identifying manipulated episodes could potentially strengthen the backdoor attack.